\documentclass[journal]{IEEEtran}

\usepackage{color,array,amsthm}
\usepackage{booktabs}
\usepackage{csvsimple} %
\usepackage{graphicx}
\usepackage{amsmath}
\usepackage{amssymb}
\usepackage[nohyperlinks]{acronym}
\usepackage{soul} %
\usepackage{subcaption}
\usepackage{svg}
\usepackage{tikz}

\newcommand{\gauss}[3]{\mathcal{N}\left(#1; \, #2, \, #3\right)}

\DeclareMathOperator*{\argmax}{arg\,max}

\begin{document}

\title{Nonlinearity and Uncertainty Informed Moment-Matching Gaussian Mixture Splitting}

\author{Jackson Kulik, Keith A.\ LeGrand\thanks{Jackson Kulik is with the Department of Mechanical and Aerospace Engineering, Utah State University, Logan, Utah, United States. Keith A.\ LeGrand is with the School of Aeronautics and Astronautics, Purdue University, West Lafayette, Indiana, United States. This work has been submitted to the IEEE for possible publication. Copyright may be transferred without notice, after which this version may no longer be accessible.}}

\markboth{KULIK \& LEGRAND}{SHORT ARTICLE TITLE}
\maketitle

\acrodef{alodt}[ALoDT]{adaptive level of detail transform}
\acrodef{cvm}[CvM]{Cram\'{e}r von Mises}
\acrodef{ekf}[EKF]{extended Kalman filter}
\acrodef{elk}[ELK]{expected likelihood kernel}
\acrodef{gm}[GM]{Gaussian mixture}
\acrodef{ise}[ISE]{integral squared error}
\acrodef{fos}[FOS]{first-order stretching}
\acrodef{lam}[LAM]{likelihood agreement measure}
\acrodef{kkt}[KKT]{Karush–Kuhn–Tucker}
\acrodef{mcr}[MCR]{maximal covariance ratio}
\acrodef{MaDEM}[MaDEM]{Mahalanobis distance of the error in the mean}
\acrodef{nise}[NISE]{normalized integral squared error}
\acrodef{nrho}[NRHO]{near rectilinear halo orbit}
\acrodef{pdf}[pdf]{probability density function}
\acrodef{sadl}[SADL]{statistical and deterministic linearization}
\acrodef{safos}[SAFOS]{spherical-average first-order stretching}
\acrodef{sasos}[SASOS]{spherical-average second-order stretching}
\acrodef{sos}[SOS]{second-order stretching}
\acrodef{solc}[SOLC]{second-order linearization change}
\acrodef{sut}[SUT]{scaled unscented transform}
\acrodef{ukf}[UKF]{unscented Kalman filter}
\acrodef{us}[US]{uncertainty-scaled}
\acrodef{usfos}[USFOS]{uncertainty-scaled first-order stretching}
\acrodef{ussolc}[USSOLC]{uncertainty-scaled second-order linearization change}
\acrodef{wsasos}[WSASOS]{whitened spherical-average second-order stretching}
\acrodef{wussadl}[WUSSADL]{whitened uncertainty-scaled statistical and deterministic linearization}
\acrodef{wusfos}[WUSFOS]{whitened uncertainty-scaled first-order stretching}
\acrodef{wussolc}[WUSSOLC]{whitened uncertainty-scaled second-order linearization change}
\acrodef{wussos}[WUSSOS]{whitened uncertainty-scaled second-order stretching}

\begin{abstract}
Many problems in navigation and tracking require increasingly accurate characterizations of the evolution of uncertainty in nonlinear systems.
Nonlinear uncertainty propagation approaches based on Gaussian mixture density approximations offer distinct advantages over sampling based methods in their computational cost and continuous representation.
State-of-the-art Gaussian mixture approaches are adaptive in that individual Gaussian mixands are selectively split into mixtures to yield better approximations of the true propagated distribution.
Despite the importance of the splitting process to accuracy and computational efficiency, relatively little work has been devoted to mixand selection and splitting direction optimization.
The first part of this work presents splitting methods that preserve the mean and covariance of the original distribution.
Then, we present and compare a number of novel heuristics for selecting the splitting direction.
The choice of splitting direction is informed by the initial uncertainty distribution, properties of the nonlinear function through which the original distribution is propagated, and a whitening based natural scaling method to avoid dependence of the splitting direction on the scaling of coordinates.
We compare these novel heuristics to existing techniques in three distinct examples involving Cartesian to polar coordinate transformation, Keplerian orbital element propagation, and uncertainty propagation in the circular restricted three-body problem.
\end{abstract}

\section{INTRODUCTION}
Many problems encountered in navigation, tracking, uncertainty quantification, and machine learning, can be distilled as the problem of mapping the uncertainty of a random variable $\mathbf{x}$ through a nonlinear function $\mathbf{g}$.
If the local behavior of $\mathbf{g}$ within the support of a Gaussian-distributed $\mathbf{x}$ is approximately linear, the mapped uncertainty is approximately Gaussian.
Further, the moments of the output Gaussian distribution are readily obtained by linearized filter approaches such as the \ac{ekf} and \ac{ukf}.
In all other cases, Gaussianity is not preserved, necessitating alternative approaches capable of fully describing the non-Gaussian output density.
One such approach is to approximate the non-Gaussian distribution as a \ac{gm}--that is, a weighted sum of smaller Gaussian distributions.
In the context of nonlinear uncertainty propagation and filtering, \ac{gm} approximations strike a balance between the computational tractability of the \ac{ekf} and \ac{ukf} and the ability to handle highly non-Gaussian distributions as in the particle filter.
Using the unscented transform \cite{julier2004unscented} or linear covariance analysis around the mean of each mixand (the covariance propagation step of the extended Kalman filter) \cite{alspach1972nonlinear}, a \ac{gm} can be propagated approximately through some nonlinear function $\mathbf{g}$.

The accuracy of \ac{gm} approaches is largely determined by how nonlinear the function is within the local effective support of each mixand.
A more granular mixture of smaller mixands--as measured by their covariance--permits better locally linear approximations than a coarse mixture of larger mixands.
On the other hand, increased mixture sizes come with additional computational cost.
For these reasons, the judicious selection of what regions of the domain require higher resolution mixture approximation is a key consideration in adaptive \ac{gm} methods.

The most common approach to increasing a \ac{gm}'s resolution is \ac{gm} splitting, wherein a selected mixand is replaced by a \ac{gm} approximation of itself.
Inherent in this approximation will be some error, though as the number of mixands increases the approximation converges to the true distribution pointwise \cite{anderson1979optimal}.
The two key considerations in \ac{gm} splitting are 1) what mixands should be split and 2) along what direction(s) should the split be performed.
Many early approaches to splitting direction selection in the aerospace community were informed strictly by the original uncertainty distribution alone \cite{demars2013entropy,vittaldev2016spacecraft}. Additionally, earlier work outside of the aerospace literature used the unscented transform to choose splitting directions along highly nonlinear directions \cite{faubel2009split, faubel2010further, leutnant2011versatile, huber2011AdaptiveGaussianMixture}.
Recently, there has been a surge of renewed interest from the aerospace community in choosing a splitting scheme based on properties of the nonlinear function \cite{jones2024physics} potentially in addition to the uncertainty distribution  \cite{tuggle2018automated, tuggle2020model, legrand2022SplitHappensImprecise}. These approaches have relied on the partial derivatives of the nonlinear function rather than the unscented transform to identify nonlinear directions. Other recent work outside of the aerospace community has chosen the splitting direction based on measures of non-Gaussianity by analyzing the skewness of the resulting distribution \cite{dunik2018directional}.
In a similar vein, recent work in the astrodynamics community has characterized the times at which dynamics are sufficiently nonlinear as characterized by tensor eigenvalue analysis to result in a distribution that is non-Gaussian %
 \cite{gutierrez2024classifying}.

In this work, we review and examine existing nonlinearity-informed methods for performing \ac{gm} splitting \cite{jones2024physics, tuggle2018automated}
in addition to proposing three novel classes of splitting methods informed by the nonlinear function.
The first novel method class relies on the theory of tensor eigenvalues and operator norms \cite{kulik2024applications,jenson2023semianalytical} to identify maximally nonlinear directions for the nonlinear function.
The second novel class of methods identifies directions in which the difference between a statistical linearization \cite{sarkka2023bayesian} and a deterministic linearization is most pronounced.
The third class incorporates averaging over a covariance ellipsoid to find directions which overall lead to strong stretching or nonlinear behavior.
We demonstrate that each of these existing and novel methods can be extended in order to account for initial uncertainty (in a manner similar to \cite{tuggle2018automated}), as well as be made independent of coordinate scaling by employing a whitening transformation.

The splitting methods developed in this work also emphasize preservation of the structure of the original uncertainty distribution--in particular the first two moments--while minimizing the approximation error in an $L_2$ sense.
Novel proofs of positive definiteness of the resulting mixand covariance and generalization for scaled homoscedastic mixands are presented in this work beyond existing similar moment-preserving approaches from Tuggle and Zanetti  \cite{tuggle2018automated, tuggle2020model}.

In order to compare and validate these splitting approaches, we test recursive splitting procedures on three nonlinear functions: the Cartesian-to-polar transformation, two-body dynamic propagation in Keplerian orbital elements, and three-body dynamic propagation in Cartesian coordinates for the nominal NASA Gateway \ac{nrho}  \cite{NationalAA2019}.
Metrics such as error in mean and covariance %
as well as likelihood agreement are employed to measure the success of each approach at approximating the true nonlinearly propagated distribution which is obtained analytically or by sampling methods.

The remainder of this work is organized as follows.
Sec.~\ref{sec:background} provides background information relevant to this work, including a review of higher-order tensors, constrained optimization, whitening transformations, and performance metrics for \ac{gm} approximation.
Moment-matched splitting is then discussed in Sec.~\ref{sec:moment_matched_splitting}.
Novel mixand selection criteria and direction optimization objectives are presented in Sec.~\ref{sec:splitting_direction}.
Three distinct applications of our methods are demonstrated in Sec.~\ref{sec:applications}.
Guidance for choosing an appropriate method is provided in Sec.~\ref{sec:discussion}, and conclusions are drawn in Sec.~\ref{sec:conclusion}.
\section{BACKGROUND AND NOTATION}
\label{sec:background}

\subsection{Tensors and Constrained Optimization}
In this work, scalars are denoted with unbolded letters, vectors by lower case bold letters, and matrices/higher-order tensors by upper case bold letters. Tensors play an important role in this work in that they allow us to describe arbitrarily high-order partial derivatives of nonlinear functions as well as higher-order statistical moments beyond second-order.
The higher-order tensors discussed here are either mixed tensors with one contravariant index and two covariant indices, or purely covariant fourth-order tensors.
The tensors that are fourth-order covariant operate on four vectors (or four times on a single vector) in order to produce a scalar.
Let $\mathbf{B}$ be such a tensor and $\mathbf{x}$ be a vector.
Then, the result of $\mathbf{B}$ acting on four vectors that are identical to $\mathbf{x}$ is denoted in shorthand and Einstein notation respectively as
\begin{equation}
  \label{eq:covariant_tensor_example}
\mathbf{B}\mathbf{x}^4=B_{i,j,k,l}x^ix^jx^kx^l
\end{equation}
Einstein notation implies a summation over each index that is repeated in an expression.
More explicitly,
\begin{equation}
    B_{i,j,k,l}x^ix^jx^kx^l=\sum_{i=1}^n\sum_{j=1}^n\sum_{k=1}^n\sum_{l=1}^n B_{i,j,k,l}x^ix^jx^kx^l
\end{equation}
where the dimension of the vector $\mathbf{x}$ is $n$ and the dimension of $\mathbf{B}$ is also $n$ along each index.
An upper index denotes contravariance whereas a lower index denotes covariance.
Vectors--represented by column vectors in matrix algebra--are contravariant whereas covectors--represented by row vectors in matrix algebra--are covariant.
Covariant portions of a tensor can act on (contract with) contravariant parts of another tensor in a manner similar to how a covector (row vector) can act on a vector (column vector) to produce a scalar under the standard inner product.
Under change of coordinates affecting the units of the coordinate system, contravariant quantities will transform in a manner proportional to the change of coordinates whereas the inverse is true for covariant quantities.
For example, a change in coordinates from [m] to [cm] will result in contravariant quantities scaling up by a factor of 100, while covariant quantities will scale down by a factor of 100.

As some additional motivation, we discuss the covariance/contravariance of two familiar second-order tensors.
The covariance matrix $\mathbf{P}$ is an example of a second-order contravariant tensor, whereas the precision matrix or inverse of the covariance matrix is a second-order covariant tensor. In matrix algebra, the expression for the squared Mahalanobis distance of a vector $\mathbf{x}$ is given by
$\mathbf{x}^T\mathbf{P}^{-1}\mathbf{x}$
which is very similar to the expression for the marginal variance $\hat{\mathbf{c}}^T \mathbf{P}\hat{\mathbf{c}}$ of a random variable with covariance $\mathbf{P}$ projected onto a given direction $\hat{\mathbf{c}}$.
However, these expressions are different in tensor notation because they signify different operations.
In the case of the marginal variance,
\begin{equation}
    \hat{c}_i\hat{c}_jP^{i,j}=\mathrm{E}(\hat{c}_i y^i\hat{c}_j y^j)=\mathrm{E}((\hat{\mathbf{c}}^T\mathbf{y})^2)
\end{equation}
if $\mathbf{y}$ has covariance $\mathbf{P}$, where $\mathrm{E}(\cdot)$ denotes the expectation operator.
Here $\hat{\mathbf{c}}^T$ is a row vector (a covariant 1-tensor) that is operating on the covariance matrix (an order 2 contravariant tensor).
In the case of the squared Mahalanobis distance $(\mathbf{P}^{-1})_{i,j}x^ix^j$, $\mathbf{x}$ is a vector (a contravariant 1-tensor) and so $\mathbf{P}^{-1}$ is a covariant tensor representing a metric.
Recall that the Mahalanobis distance is invariant to linear changes of coordinates.
Thus, as $\mathbf{x}$ is rescaled, the precision matrix must be rescaled in the opposite sense to keep the Mahalanobis distance the same in either coordinate system.
This offers more insight into  why it matters that the precision matrix is treated as a covariant tensor and the covariance matrix as a contravariant tensor.

Mixed higher-order tensors, which are neither purely covariant nor purely contravariant, also play a role in this work.
A tensor $\mathbf{C}$ with one contravariant index and two covariant indices acts on two vectors to produce a vector.
In the case where the tensor acts on the same vector twice, the $i$th entry of the resulting vector is given by
\begin{equation}
  \label{eq:mixed_tensor_example}
    (\mathbf{C}\mathbf{x}^2)^i=C^i_{j,k}x^jx^k
\end{equation}
where the lack of two $i$ indices in each expression indicates that no summation is conducted along that axis.
The order of the resulting tensor is given by the number of such uncontracted indices.
As seen in both~\eqref{eq:covariant_tensor_example} and~\eqref{eq:mixed_tensor_example}, we utilize a shorthand where tensors placed next to one another indicate a contraction along some natural pairing of indices.
A power placed on a vector next to a tensor indicates contraction of the tensor with that many copies of the vector.
When the pairing of indices is not clear, the expression will also be written in Einstein notation.

In this work, we will often be concerned with the optimization of the norm of a matrix vector product or tensor vector double contraction subject to a norm constraint on the vector.
In the matrix case, if both the objective function and constraint are 2-norms, the constrained optimization problem is solved by the right singular value of the matrix.
Furthermore, the maximum is given by the induced 2-norm of the matrix or its largest singular value:
\begin{equation}
  \Vert\mathbf{A}\Vert_2=\max_{\Vert\mathbf{x}\Vert_2=1}\Vert\mathbf{A}\mathbf{x}\Vert_2=\sigma_{\mathrm{max}}(\mathbf{A})
\end{equation}
as shown in \cite[Sec.~2.3]{golub2013matrix}.

Other induced norms of matrices prove useful in the development of our splitting heuristics.
For example, constraint can be based on a norm induced by some quadratic form (covariant second-order tensor) such as the precision matrix:
\begin{equation}
    \Vert \mathbf{x}\Vert_{\mathbf{P}^{-1}}^2=\mathbf{x}^T\mathbf{P}^{-1}\mathbf{x}
    \label{eq:mh_norm}
\end{equation}
Geometrically, an equality constraint on the norm induced by some quadratic form corresponds to the surface of an ellipsoid.
The semi-axes of the ellipsoid are determined by the square roots of the eigenvalues multiplied by the eigenvectors of the matrix.
The norm induced by the precision matrix in~\eqref{eq:mh_norm} is the Mahalanobis distance.
The ellipsoids corresponding to a constraint on the Mahalanobis distance are so-called covariance ellipsoids and represent constant probability surfaces in the case of the Gaussian distribution.

These more general constrained optimizations of the form
\begin{equation}
    \max_{\mathbf{x}^T\mathbf{B}\mathbf{x}=1}\Vert\mathbf{A}\mathbf{x}\Vert_2
\end{equation}
can be solved using a change of variables and a singular value decomposition.
The optimization can also be performed by solving a generalized eigenvalue problem as is discussed for particular cases in Sec.~\ref{sec:splitting_direction}.\ref{sec:hybrid_methodologies}.
In the tensor case, similar constrained optimization problems can be posed as induced tensor norms.
Then, solutions may be obtained numerically by solving tensor generalizations of eigenvalue and singular value problems \cite{kulik2024applications, jenson2023semianalytical}.
We briefly summarize some of this theory in Sec.~\ref{sec:splitting_direction}.\ref{sec:stretching_informed_heuristics}.\ref{sec:second-order}.

\subsection{Whitening Transformations}
Having discussed the Mahalanobis distance above we can introduce the related notion of a whitening transformation. A whitening transformation takes covariance ellipsoids to spheres so that the transformed vector is distributed with covariance equal to the identity matrix. There are many whitening transformations corresponding to different matrix square roots of the original covariance and different rotations of the sphere that was output. If a matrix square root $\mathbf{P}^{1/2}$ is defined such that
\begin{equation}
    \mathbf{P}=\mathbf{P}^{1/2}(\mathbf{P}^{1/2})^T
\end{equation}
then the inverse of the matrix square root is a whitening transformation
\begin{equation}
    \mathbf{y}=\mathbf{P}^{-1/2}\mathbf{x}
\end{equation}
because
\begin{equation}
    \mathrm{E}(\mathbf{y}\mathbf{y}^T)=\mathbf{I}
\end{equation}
The standard 2-norm in a whitened space is the Mahalanobis distance in the original space:
\begin{equation}
  \sqrt{\mathbf{y}^T\mathbf{y}}
  =
  \sqrt{\mathbf{x}^T\mathbf{P}^{-1}\mathbf{x}}
\end{equation}
The Mahalanobis distance can express quantities in units that are easily understood with respect to the scale of the uncertainty. The expected squared Mahalanobis distance is given by $m$, the dimension of $\mathbf{z}$, which is the expectation of a Chi-square random variable with $m$ degrees of freedom. Thus, the squared Mahalanobis distance of a vector can be compared to its dimension to determine whether or not the vector is large relative to the uncertainty associated with the covariance that induces that Mahalanobis distance.

\section{MOMENT MATCHING SPLITTING}
\label{sec:moment_matched_splitting}
We present a method for performing Gaussian mixture splitting in a manner that preserves the mean and covariance of the original distribution.
This generalizes an existing methodology that requires homoscedasticity \cite{tuggle2018automated}.

\subsection{UNIVARIATE SPLITTING}
Splitting can be performed efficiently by leveraging a pre-generated library of optimal solutions for the univariate standard Gaussian $q(x)$, as first proposed by Hanebeck in \cite{hanebeck2003ProgressiveBayesNew}.
Once found, optimal univariate solutions are quickly referenced at runtime and mapped to multivariate split solutions using the method described in the following subsection.

Let
\begin{align}
    \tilde{q}(x) = \sum_{i=1}^{L} \tilde{w}_{j}
    \gauss{x}{\tilde{\mu}_{i}}{\tilde{\sigma}_{i}^{2}}
\end{align}
represent the univariate mixture approximation of $q(x)$.
Then, the univariate split library is populated by the solutions of
\begin{align}
\label{eq:univariate_split_objective}
    \min& \left(L_{2}(q||\tilde{q}) + \frac{\lambda}{L} \sum_{i=1}^{L}\tilde{\sigma}_{i}^{2}\right)\\
    \textrm{s.t.}& \sum_{i=1}^{L}\tilde{w}_{i} = 1\\
    & \tilde{w}_{i} > 0 \,\, \forall\,\, i =1,\hdots, L\\
    & \operatorname{var}_{\tilde{q}}(x) = 1
\end{align}
over a range of desired mixture sizes $L$ and parameters $\lambda$.
The regularization term $\lambda$ promotes smaller mixand variances and larger mixand mean separation and avoids the trivial solutions of $\tilde{\mu}_{i}=0$, $\tilde{\sigma}_{i}=1$.
While any distance measure can be used in~\eqref{eq:univariate_split_objective}, the $L_{2}$ distance (a.k.a. \ac{ise}) is appealing because it is available in closed form for \ac{gm} inputs.
To reduce the optimization problem dimension, we follow  \cite{demars2013entropy} and restrict the solution space to homoscedastic mixtures ($\tilde{\sigma}_{i}=\tilde{\sigma}$) with equally spaced means
\begin{align}
    \tilde{\mu}_{i} = \varepsilon L \left(
    \frac{i-1}{L-1} - \frac{1}{2}
    \right)
\end{align}
where $\varepsilon$ is the mean spacing.
The common mixand variance that preserves the original distribution variance is determined by the weight and mean selections and given by
\begin{align}
(\tilde{\sigma}^{*})^2 = 1 - \sum_{i=1}^{L} \tilde{w}_{i} \tilde{\mu}_{i}^{2}
\end{align}
By directly enforcing variance preservation, the mixand variance is no longer a free parameter, and the feasibility constraint
\begin{align}
    & \sum_{i=1}^{L} \tilde{w}_{i} \tilde{\mu}_{i}^2 < 1
\end{align}
is imposed to guarantee positive definite mixand variances.
While homoscedasticity simplifies the univariate library solution, this assumption is not required by the multivariate splitting method presented in the following subsection.

\subsection{MULTIVARIATE SPLITTING}
\label{sec:multivariate_splitting}
Given a splitting direction and univariate splitting scheme that preserves the mean and variance along that direction, we present a method to select covariances for each mixand such that the overall mixture mean and covariance are preserved.
This method will choose mixand covariances that are identical up to a scaling factor.
This scaling factor corresponds to the ratio of the variance of each mixand in the univariate splitting being employed.
This method generalizes the independent work of both \cite{leutnant2011versatile,tuggle2020model} which also presented multivariate moment matching splittings along arbitrary directions.
Both existing methods operated under the assumption of homoscedastic mixand covariances rather than mixand covariances that are allowed to be scalar multiples of one another.
We also present a novel proof of the resulting positive definiteness of each mixand covariance where the existing proof from \cite{tuggle2020model} focuses on the positivity of the variance along the splitting direction.

Assume the univariate splitting in the direction $\hat{\mathbf{x}}^*$ has $n$ mixands and is given by $(w_i,\mu_i, \sigma_i)$.%
We choose each mixand to have covariance $\sigma_i^2 \bar{\mathbf{P}}$.
The covariance of the mixture distribution can be written
\begin{equation}
    \mathbf{P}=\sum_{i=1}^L w_i\left[(\boldsymbol{\mu}_i-\boldsymbol{\mu})(\boldsymbol{\mu}_i-\boldsymbol{\mu})^T+\sigma_i^2 \bar{\mathbf{P}}\right]
    \label{eq:mixture-cov}
\end{equation}
where the mixand means are placed along the splitting direction $\hat{\mathbf{x}}^*$ relative to the original mean as
\begin{equation}
    \boldsymbol{\mu}_i=\boldsymbol{\mu}+\mu_i\hat{\mathbf{x}}^*
    \label{eq:mean_placement}
\end{equation}
Rearranging~\eqref{eq:mixture-cov}, we see that the covariance matching mixand covariance is given by a rank-1 downdate of the original covariance (since the mixand means all fall into a line so that the sum of the outer products is still a rank-1 matrix)
\begin{equation}
    \bar{\mathbf{P}}=\frac{\mathbf{P}-\sum_{i=1}^L w_i(\boldsymbol{\mu}_i-\boldsymbol{\mu})(\boldsymbol{\mu}_i-\boldsymbol{\mu})^T}{\sum_{i=1}^{L} w_i\sigma_i^2}
    \label{eq:mixand-cov}
\end{equation}

While no known analytical methods are available to perform a rank-1 update of the eigenvalue decomposition, fast numerical methods \cite{bunch1978rank} may be used.
In general, the eigendecomposition--and thus the elliptical geometry associated with the covariance matrix--of the children mixands changes in a non-trivial manner with respect to the parent mixand.
The exception to this fact is when the splitting direction is aligned with an eigenvector of the original covariance, a principal axis of the original covariance ellipsoid.
If the \ac{gm} is being employed alongside some square-root algorithm, it is important to note that there exist fast rank-1 downdate algorithms for the Cholesky decomposition that can make it very efficient to compute the Cholesky factors of the mixand covariances \cite{bojanczyk1987note}.
For use in a rank-1 downdate, we note that the rank-1 matrix in question can be written explicitly as the outer product

\begin{align}
   &\sum_{i=1}^{L} w_i(\boldsymbol{\mu}_i-\boldsymbol{\mu})(\boldsymbol{\mu}_i-\boldsymbol{\mu})^T =\\&\left(\sqrt{\sum_{i=1}^{L} w_i\Vert\boldsymbol{\mu}_i-\boldsymbol{\mu}\Vert^2}\hat{\mathbf{x}}^*\right)\left(\sqrt{\sum_{i=1}^{L} w_i\Vert\boldsymbol{\mu}_i-\boldsymbol{\mu}\Vert^2}\hat{\mathbf{x}}^*\right)^T\nonumber
\end{align}
where $\hat{\mathbf{x}}^*$ is the unit vector along which the means of the mixands are distributed, or explicitly
\begin{equation}
    \hat{\mathbf{x}}^*=\frac{\boldsymbol{\mu}_j-\boldsymbol{\mu}}{\Vert \boldsymbol{\mu}_j-\boldsymbol{\mu}\Vert_2}
\end{equation}
for any choice of $j$ such that $\boldsymbol{\mu}_j\neq\boldsymbol{\mu}$.

While we have given an expression for the covariance of the mixands under a given splitting scheme, we have not determined whether the resulting expression does, in fact, yield a valid covariance.
The question of whether a given mixture configuration $(w_i,\mu_i, \sigma_i)$ yields a possible covariance matching homoscedastic (up to a scaling) mixture depends on whether $\bar{\mathbf{P}}$ is positive semi-definite.
We prove conditions for positive definiteness of the mixand covariances that result from the multivariate moment preserving splitting scheme in Sec. \ref{sec:posdef_app}. We summarize the most important takeaways here. The mixand covariances will be positive semi-definite if and only if the following inequality holds in terms of the original covariance along with the mixand weights and means.
\begin{equation}
    \label{eq:spd-cond-text}
    \sum_{i=1}^{L} w_i\Vert\boldsymbol{\mu}_i-\boldsymbol{\mu}\Vert^2 \leq \frac{1}{(\hat{\mathbf{x}}^*)^T\mathbf{P}^{-1}\hat{\mathbf{x}}^*}
\end{equation}
In the case of a \textit{univariate} Gaussian, the variance and the reciprocal of the precision are identical.
While it seems natural to design a univariate splitting of a \textit{multivariate} Gaussian using the variance along the splitting direction, the positive definiteness condition in~\eqref{eq:spd-cond-text} implies that the reciprocal precision along the splitting direction is actually the parameter relevant for designing the splitting.
A valid (having non-negative mixand variances) univariate splitting library that preserves reciprocal of precision working in tandem with choice of mixand covariance from~\eqref{eq:mixand-cov} will guarantee that the mixand covariance is symmetric positive semi-definite as required. If the directional variance rather than the directional precision were employed for the univariate splitting it is possible that the resulting mixand covariances would not be positive semi-definite.
That is, the mixture parameters $(w_i,\mu_i, \sigma_i)$ should be designed for splitting the univariate distribution with mean zero and variance
\begin{equation}
\sigma^2_{\hat{\mathbf{x}}}=\left((\hat{\mathbf{x}}^*)^T\mathbf{P}^{-1}\hat{\mathbf{x}}^*\right)^{-1}
\label{eq:reciprocal_precision}
\end{equation}
This choice of directional reciprocal precision was also previously derived as the conditional directional variance conditioned on all other directions  \cite{tuggle2018automated}.
Using a univariate splitting library based on this univariate variance of the original distribution, and then constructing the mixand covariance according to~\eqref{eq:mixand-cov} is guaranteed to result in valid mixands with positive semi-definite covariances.

To summarize, given a splitting direction $\hat{\mathbf{x}}^*$, the reciprocal precision along that direction is calculated according to \eqref{eq:reciprocal_precision}, then a univariate Gaussian mixture is chosen from a library to split a univariate mean zero Gaussian with variance given by the reciprocal precision from \eqref{eq:reciprocal_precision} into a univariate Gaussian mixture with parameters $(w_i,\mu_i, \sigma_i)$.
Then, the final multivariate Gaussian mixture will have the same weights $w_i$, means $\boldsymbol{\mu}_i$ given by \eqref{eq:mean_placement}, and covariances given by
\begin{equation}
    \mathbf{P}_i=\sigma_i^2\frac{\mathbf{P}-\sum_{i=1}^L w_i(\boldsymbol{\mu}_i-\boldsymbol{\mu})(\boldsymbol{\mu}_i-\boldsymbol{\mu})^T}{\sum_{i=1}^{L} w_i\sigma_i^2}
\end{equation}
where $\boldsymbol{\mu}$ and $\mathbf{P}$ are the mean and covariance, respectively, of the original multivariate Gaussian being split.

\section{SPLITTING DIRECTION AND CRITERION}
\label{sec:splitting_direction}
Given a random variable $\mathbf{x}\sim \mathcal{N}(\boldsymbol{\mu}_x, \mathbf{P}_x)$ and a nonlinear function $\mathbf{g}: \mathbb{R}^{n}\rightarrow\mathbb{R}^m$, we consider methods to determine along what direction the original Gaussian distribution should be split.
The overarching objective is that the resulting mixture $(w^{(i)}, \boldsymbol{\mu}^{(i)}_x,\mathbf{P}^{(i)}_x)$, when propagated through $\mathbf{g}$ using linear covariance techniques, forms a good approximation of the distribution of $\mathbf{g}(\mathbf{x})$ described by the density $p(\mathbf{z})$.

Early techniques for \ac{gm} splitting chose to split along directions of maximal uncertainty  \cite{demars2013entropy, cmes}.
Recently, there is emerging interest in choosing the splitting based off of properties of the map $\mathbf{g}$--in particular, stretching properties of the linearization \cite{jones2024physics, calkins2024} or how that linearization changes with the reference point for linearization \cite{tuggle2018automated, tuggle2020model}.
Sec.~\ref{sec:splitting_direction}.\ref{sec:stretching_informed_heuristics} reviews existing heuristics and introduces new heuristics pertaining to the map.
Sec.~\ref{sec:splitting_direction}.\ref{sec:uncertainty} then combines these heuristics with uncertainty based approaches.
Sec.~\ref{sec:splitting_direction}.\ref{sec:output_whitening} develops a novel whitening based method to render these heuristics invariant under rescalings of coordinates while providing meaningful splitting criteria.
Sec.~\ref{sec:splitting_direction}.\ref{sec:output_whitening} introduces novel heuristics based on spherical averages, and hybrid methods based on the aforementioned techniques are discussed in Sec.~\ref{sec:splitting_direction}.\ref{sec:hybrid_methodologies}.

\subsection{STRETCHING INFORMED HEURISTICS}
\label{sec:stretching_informed_heuristics}
Each method for choosing the splitting direction based on properties of the nonlinear function $\mathbf{g}$ is phrased in terms of some constrained optimization problem of the form
\begin{equation}
    \hat{\mathbf{x}}^*=\argmax_{\Vert\mathbf{x}\Vert_2=1}\mathcal{F}(\mathbf{g},\mathbf{x})
    \label{eq:unit_constrained_opt}
\end{equation}
where $\mathcal{F}$ is some functional acting on functions $\mathbf{g}$ and vectors $\mathbf{x}$ to produce a non-negative real number.

\subsubsection{LINEAR STRETCHING DIRECTIONS}
The most basic method for choosing a function-informed splitting direction is to use the maximal linear stretching direction of the function $\mathbf{g}$ linearized at the mean $\boldsymbol{\mu}_x$. Let the Jacobian of the nonlinear function $\mathbf{g}$ be defined as
\begin{equation}
    \mathbf{G}=\frac{\partial\mathbf{g}}{\partial \mathbf{x}}\big\vert_{\mathbf{x}=\boldsymbol{\mu}_x}
\end{equation}
where the functional dependence of $\mathbf{G}$ on $\boldsymbol{\mu}_{x}$ is omitted for brevity.
Then, the functional for optimization is
\begin{equation}
  \label{eq:first_order_stretching_objective}
    \mathcal{F}=\left\Vert\mathbf{G}\mathbf{x}\right\Vert_2
\end{equation}
Then, the optimal splitting direction $\hat{\mathbf{x}}^*$ is the right singular vector corresponding to the largest singular value of $\mathbf{G}$.
Alternatively, the splitting direction can be obtained as the eigenvector corresponding to the maximal eigenvalue of the ``square" of the Jacobian $\mathbf{G}^T\mathbf{G}$.
In the special case where $\mathbf{g}$ represents the flow of a dynamical system, $\mathbf{G}^{T}\mathbf{G}$ is known as the Cauchy-Green strain tensor.

The underlying premise of~\eqref{eq:first_order_stretching_objective} as a splitting heuristic is that a distribution propagated through a nonlinear function is often more poorly approximated by linear covariance analysis far from the mean.
By this logic, directions in the input space that get stretched the most in the output space may be good choices to split along.
This method seems to work well in long-term two- and three-body astrodynamics applications  \cite{jones2024physics} while faring poorly in the entry, descent, and landing \cite{calkins2024}, as well as in the context of measurement functions as will be discussed in Sec.~\ref{sec:applications}.
We refer to this method of selecting the splitting direction as \ac{fos}.
\subsubsection{SECOND-ORDER STRETCHING DIRECTIONS}
The \ac{fos} approach promotes splitting up the distribution along the directions that will be stretched the most up to a linear approximation.
However, since linear functions preserve Gaussianity, we might place more importance on input directions along which the function is most nonlinear, or along which the linear approximation is most inaccurate.
These directions along which the linear approximation is most inaccurate will lead to deviations from Gaussianity in the resulting distribution.
This approach has been taken using techniques from differential algebra \cite{losacco2024low} to identify and split along Cartesian coordinate axis-aligned directions along which the nonlinear contributions of that coordinate are above some threshold.
Coincidentally, in the context of a flow of a dynamical system in an astrodynamics setting, the input direction of maximal linear stretching tends to coincide very closely with the input direction of maximal nonlinearity \cite{boone2023directional}.
Thus, the heuristic based on maximal linear stretching directions tends to coincide well with the direction of maximal nonlinearity for two- and three-body dynamics and may explain its success in those cases \cite{jones2024physics} but failure in the case of entry, descent, and landing dynamics \cite{calkins2024}.

One definition of nonlinear directions of maximal stretching can be calculated using higher-order versions of the Cauchy-Green strain tensors which arise in the Taylor expansion of the 2-norm of a perturbation of the flow map (or more generally a nonlinear function) \cite{jenson2023semianalytical,jenson2024bounding}.
In our context, we will define maximal error stretching directions as the direction that gives the maximal 2-norm of the second-order contribution from the function $\mathbf{g}$.
That is, given the nonlinear function $\mathbf{g}$, we examine the second-order partial derivative tensor written as a whole or in coordinates as
\begin{align}
    \mathbf{G}^{(2)}&=\frac{\partial^2\mathbf{g}}{\partial\mathbf{x}^2} \big|_{\mathbf{x}=\boldsymbol{\mu}_{x}}
\\
    G^i_{j,k}&=\frac{\partial g^i}{\partial x^j\partial x^k} \big|_{\mathbf{x}=\boldsymbol{\mu}_{x}}\end{align}
Second-order partial derivatives of the flow of a dynamical system can be found using variational equations \cite{park2006nonlinear} or differential algebra techniques \cite{rasotto2016differential} which give rise to the second-order state transition tensor.
Differential algebra or analytical techniques can be used to find the second-order partial derivatives for nonlinear measurement functions.

The maximal second-order error stretching direction will be given by the unit vector that solves the following constrained optimization problem from~\eqref{eq:unit_constrained_opt} with the functional
\begin{equation}
    \mathcal{F}=\Vert\mathbf{G}^{(2)}\mathbf{x}^2\Vert_2
    \label{eq:max_nonlin}
\end{equation}
where we use shorthand for the tensor vector multiplication that gives rise to a vector with components
\begin{equation}
    (\mathbf{G}^{(2)}\mathbf{x}^2)^i=G^i_{j,k}x^jx^k
\end{equation}
This constrained optimization problem can be reformulated into a tensor Z-eigenvalue problem in terms of the square of the tensor by the method of Lagrange multipliers:
\begin{align}
    \Tilde{\mathbf{G}}^{(2)}\mathbf{x}^4=\lambda \mathbf{x}\\
    \Vert\mathbf{x}\Vert_2=1
\end{align}
where $\Tilde{\mathbf{G}}^{(2)}$ is the ``square" of the tensor $\mathbf{G}^{(2)}$, defined as
\begin{equation}
    (\Tilde{\mathbf{G}}^{(2)})_{i,j,k,l}=(\mathbf{G}^{(2)})^p_{i,j}\delta_{p,q}(\mathbf{G}^{(2)})^q_{k,l}
\end{equation}
with the metric tensor for Euclidean space given by the Kronecker delta $\delta_{p,q}$, which is unity when $p=q$ and zero when $p\neq q$.

This tensor eigenvalue problem can then be solved by shifted symmetric higher-order power iteration to find the eigenvector associated with the largest eigenvalue  \cite{kolda2011shifted, qi2005eigenvalues}. The eigenvectors and the square root of the eigenvalue give the solution to the constrained maximization problem from~\eqref{eq:unit_constrained_opt} and~\eqref{eq:max_nonlin}. Shifted symmetric higher-order power iteration is given by the update rule
\begin{equation}
    \mathbf{x}_{(k+1)}=\frac{\Tilde{\mathbf{G}}^{(2)}\mathbf{x}_{(k)}^3+\eta \mathbf{x}_{(k)}}{\Vert\Tilde{\mathbf{G}}^{(2)}\mathbf{x}_{(k)}^3+\eta \mathbf{x}_{(k)}\Vert_2}
\end{equation}
with the shorthand of a fourth-order tensor contracted with three vectors defined in index notation as
\begin{equation}
    (\Tilde{\mathbf{G}}^{(2)}\mathbf{x}_{(k)}^3)^i=\delta^{i,p}(\Tilde{\mathbf{G}}^{(2)})_{p,q,r,s}x_{(k)}^q x_{(k)}^r x_{(k)}^s
\end{equation}
where  $\delta^{i,p}$ is the inverse metric tensor and the shift $\eta$ is chosen conservatively to guarantee convergence as the sum of absolute values of the entries
\begin{equation}
  \label{eq:shift_parameter}
    \eta=\sum_{i,j,k,l} \vert (\Tilde{\mathbf{G}}^{(2)})_{i,j,k,l}\vert
\end{equation}
An initial guess must be chosen.
Shifted symmetric higher-order power iteration converges globally from any starting condition, though not always to the maximal eigenvalue/eigenvector pair as is the case with the matrix power iteration algorithm.
Instead, convergence to one of the larger eigenpairs is typically observed with the gap between the largest eigenvalue and the next largest eigenvalue playing a role in determining the convergence basin for the largest eigenpair.
In the case of astrodynamics applications, the maximal linear stretching direction makes a good initial guess for shifted symmetric higher-order power iteration  \cite{boone2023directional}, but in other instances, multiple initial guesses may need to be employed to find the maximal eigenpair.
It should be noted that the tensor $\Tilde{\mathbf{G}}^{(2)}$ is not symmetric as constructed, but that shifted symmetric higher-order power iteration works identically on this tensor as it would if the tensor were explicitly symmetrized.
This is because this tensor is symmetric enough in some sense \cite{kulik2024applications}. To be precise, for every index of the tensor there always exists a permutation of the indices that moves that chosen index into the first position and leaves the tensor unchanged. Additionally, the first index is always left uncontracted by the power iteration algorithm. As such, every permutation of the tensor yields the same result from power iteration and the original tensor will give the same result as the symmetrized tensor which is just the average over all permutations.
While more aggressive shifts are available that improve the rate of convergence, the shift proposed in \eqref{eq:shift_parameter} is convenient because it does not require tensor symmetrization.
Alternatively, the symmetrization of the tensor can be used in computing $\eta$ to facilitate faster convergence.%

While the solution to the constrained optimization problem using ~\eqref{eq:max_nonlin} corresponds to the maximum eigenvalue and eigenvector pair for a tensor Z-eigenvalue problem, the other eigenvalue and eigenvector pairs represent other \ac{kkt} points associated with the constrained optimization using~\eqref{eq:max_nonlin}.
However, unlike the right singular vectors which arise as \ac{kkt} points for a similar optimization with a matrix rather than a higher-order tensor, these \ac{kkt} points do not have the property of orthogonality that right singular vectors do.
So, while singular vector analysis of the Jacobian gives a potential orthogonal grid splitting of multiple best splitting directions in a certain sense, the \ac{kkt} points associated with the second-order partial derivative tensor do not have the requisite orthogonality to perform splitting along multiple directions simultaneously.
Instead, recursive splitting is necessary in order to split along multiple directions.
If the splitting is still relatively localized, the optimal splitting direction according to the second-order derivative tensor will not vary much between recursive splitting steps, which could be problematic.
One method to ensure splitting will happen along a variety of directions that are highly nonlinear is to recursively split with a splitting heuristic that takes into account nonlinearity and uncertainty, since uncertainty along highly nonlinear directions should be reduced at each splitting, promoting splits in alternative directions at the next recursion.
This motivates incorporating uncertainty into our splitting direction computation as discussed in Sec.~\ref{sec:splitting_direction}.\ref{sec:uncertainty}. We refer to the second order nonlinear based method of selecting the splitting direction discussed in this section without accounting for the original level of uncertainty in each direction as \ac{sos}.
\subsubsection{CHANGE IN LINEARIZATION}
The methods described in the previous section capture the error in linear covariance analysis by examining the most nonlinear directions of the function $\mathbf{g}$.
However, these methods rely on potentially repeated application of shifted symmetric higher-order power iteration for computation if a good initial guess is not know a priori.
Another method that has been presented in the literature that characterizes maximally nonlinear directions of the function $\mathbf{g}$ but exploits computational methods from linear algebra was derived by Tuggle and Zanetti  \cite{tuggle2018automated,tuggle2020model}.
Their method additionally took into account uncertainty as well in a manner that will be discussed in Sec.~\ref{sec:splitting_direction}.\ref{sec:uncertainty}.
Initially, we present a slightly different form that is mathematically identical to the nonlinearity measure from their work under the assumption of an isotropic initial distribution.
We characterize the Frobenius norm of the change in the Jacobian evaluated at a point other than the mean as approximated with the second-order partial derivative tensor. That is
\begin{equation}
    \mathcal{F}=\Vert\mathbf{G}^{(2)}\mathbf{x}\Vert_F=\Vert\bar{\mathbf{G}}^{(2)}\mathbf{x}\Vert_2
    \label{eq:tuggle}
\end{equation}
where the change in the Jacobian as the reference point moves from the mean $\boldsymbol{\mu}_x$ by $\mathbf{x}$ is approximately given by the matrix $\delta\mathbf{G}$ with entries
\begin{equation}
    (\delta\mathbf{G})^i_j=(\mathbf{G}^{(2)}\mathbf{x})^i_j=(\mathbf{G}^{(2)}\mathbf{x})^i_{j,k}x^k
\end{equation}
and the matricized or flattened tensor $\bar{\mathbf{G}}^{(2)}$ is given by the $mn$ by $n$ dimension matrix
\begin{equation}
    (\bar{\mathbf{G}}^{(2)})^{ni+j}_k=(\mathbf{G}^{(2)})^i_{j,k}
\end{equation}
Thus, the constrained optimization of~\eqref{eq:tuggle} according to~\eqref{eq:unit_constrained_opt} is given by the dominant right singular vector of the matrix $\bar{\mathbf{G}}^{(2)}$. We refer to this method of selecting the splitting direction as \ac{solc} method.

\subsubsection{STATISTICAL AND DETERMINISTIC LINEARIZATION DIFFERENCING}
The previous two methods rely on computation of higher-order derivative tensors which can be costly and unintuitive for those unused to working in this framework. Sigma point methods offer an advantage in that the evaluation of the nonlinear transformation $\mathbf{g}(\mathbf{x})$ at the sigma points (or more generally, regression points) can be compared to the linearized system to assess nonlinearity without having to directly calculate second-order derivatives. Consider the statistical linearization
\begin{align}
  \mathbf{z} = \mathbf{g}(\mathbf{x}) \approx \mathbf{G}^{(\textrm{SL})} \mathbf{x} + \mathbf{b}
\end{align}
The matrix $\mathbf{G}^{(\textrm{SL})}$ and vector $\mathbf{b}$ that minimize the mean squared error for this affine model over some sampled points are
\begin{align}
  \label{eq:statistical_linearization_matrix}
\mathbf{G}^{(\textrm{SL})}=\left(\mathbf{P}_{x z}\right)^{T}\left(\mathbf{P}_x\right)^{-1} \text { and } \mathbf{b}=\boldsymbol{\mu}_z-\mathbf{G}^{(\textrm{SL})} \boldsymbol{\mu}_x
\end{align}
where the output mean $\boldsymbol{\mu}_{z}$ and cross-covariance $\mathbf{P}_{xz}$ (as well as the covariance $\mathbf{P}_{z}$) are readily obtained via the unscented transform or other sigma point transformation  \cite{huber2011AdaptiveGaussianMixture}.
The linearization error is given by
\begin{align}
    \mathbf{e} = \mathbf{g}(\mathbf{x}) - (\mathbf{G}^{(\textrm{SL})}\mathbf{x} + \mathbf{b})
\end{align}
and is zero-mean with covariance
\begin{align}
    \mathbf{P}_{e} =\mathbf{P}_{z}-\mathbf{G}^{(\textrm{SL})} \mathbf{P}_{x} \left(\mathbf{G}^{(\textrm{SL})}\right)^{T}
    \label{eq:error_covariance}
\end{align}
Given the true Jacobian $\mathbf{G}$ of the nonlinear function $\mathbf{g}$ evaluated at the mean and the statistical linearization $\mathbf{G}^{(\textrm{SL})}$, another interesting quantity is their difference.
The right singular vectors of the difference in these two matrices may give promising directions in which the statistical linearization is substantially different from the true linearization due to strong nonlinear effects.
The primary direction where these two linearizations differ is obtained by solving the constrained optimization with objective function
\begin{equation}
  \mathcal{F}=\Vert(\mathbf{G}^{(\textrm{SL})}-\mathbf{G})\mathbf{x}\Vert_2
\end{equation}
We refer to this method of selecting the splitting direction as \ac{sadl}.
\subsection{UNCERTAINTY INFORMED HEURISTICS}
\label{sec:uncertainty}
While considering the behavior of the nonlinear function $\mathbf{g}$ is important, accounting for non-isotropic uncertainty is also important.
In the extreme case, if a multivariate Gaussian has no uncertainty at all in the direction informed by the nonlinear function, then splitting in that direction will have no utility.
To that end, the uncertainty along a candidate direction should be considered in concert with the nonlinear function properties when selecting a splitting direction.
This can be achieved in a straightforward manner by maintaining the aforementioned objectives $\mathcal{F}$ but imposing a different, uncertainty informed constraint.
Consider the following optimization
\begin{equation}
    \hat{\mathbf{x}}^*\sim\argmax_{\mathbf{x}^T\mathbf{P}_x^{-1}\mathbf{x}=1}\mathcal{F}(\mathbf{g},\mathbf{x})
    \label{eq:ellipsoid_constrained_opt}
\end{equation}
Under the constraint in this optimization, deviations from the mean are chosen to lie on the 1-sigma contour of the initial distribution.
This optimization yields the following interpretations for each of the previous objective functions:
\begin{enumerate}
    \item What direction leads to the most uncertain direction in the output space under a linear approximation?
    \item When mapping any covariance ellipsoid associated with $\mathbf{P}_x$ through the nonlinear function $\mathbf{g}$ and its linearization $\mathbf{G}$, what element of that ellipsoidal set leads to the greatest error between the two outputs (approximated up to second-order)?
    \item What point on a given covariance ellipsoid leads to the largest change in the Jacobian as compared with the Jacobian at the mean (measured in terms of the Frobenius norm)?
    \item What is the largest difference in how the deterministic linearization and statistical linearization map any point on the 1-sigma covariance ellipsoid?
\end{enumerate}
Conveniently, the optimization in~\eqref{eq:ellipsoid_constrained_opt} can be carried out over the unit ball in order to preserve the previous computational methods by applying the following change of coordinates to the input space
\begin{equation}
    \mathbf{y}=\mathbf{P}_x^{-1/2}\mathbf{x}
\end{equation}
where $\mathbf{P}^{1/2}$ is some matrix square root of the initial covariance
\begin{equation}
    \mathbf{P}_x=\mathbf{P}^{1/2}\mathbf{P}^{T/2}
\end{equation}
such as the Cholesky factor, and $\mathbf{P}_x^{-1/2},\mathbf{P}_x^{T/2}$ denotes its inverse and transpose respectively.
This coordinate transformation results in the optimization
\begin{equation}
    \hat{\mathbf{x}}^*\sim \mathbf{P}_x^{1/2}\argmax_{\mathbf{y}^T\mathbf{y}=1}\mathcal{F}(\mathbf{g},\mathbf{P}_x^{1/2}\mathbf{y})
    \label{eq:sphere_ellipsoid_constrained_opt}
\end{equation}
For each objective function, the $\mathbf{P}_x^{1/2}$ term can be folded into the matrices or tensors of the optimization explictly, or used implictly as a part of the computation of the optimization problem. This involves solving a generalized symmetric eigenvalue problem instead of the singular value decomposition in the matrix cases and as a backpropagation like step in shifted symmetric higher-order power iteration for the tensor optimizations. Note that by combining the functional from~\eqref{eq:tuggle} and the optimization from~\eqref{eq:ellipsoid_constrained_opt} or~\eqref{eq:sphere_ellipsoid_constrained_opt} we obtain a mathematically equivalent optimization to that presented by Tuggle and Zanetti \cite{tuggle2018automated}.
We will refer to methods that included initial uncertainty scaling in the optimization process for splitting direction selection as \ac{us}.
Another method for incorporating uncertainty into splitting direction selection involves simply multiplying the given direction that does not take into account initial uncertainty with the initial covariance matrix.
Choosing the resulting direction for splitting minimizes the final mixand uncertainty along the originally specified direction \cite{kumar2022splitting}.
However, this method assumes the multivariate splitting scheme from \cite{demars2013entropy, vittaldev2016spacecraft} that does not preserve the covariance of the overall mixture.
As such, we focus on uncertainty weighting based approaches described above which can be used in tandem with covariance preserving splitting methods.
\subsection{OUTPUT WHITENING-BASED HEURISTICS}
\label{sec:output_whitening}
Through uncertainty scaling, the aforementioned heuristics account for both the initial uncertainty and properties of the nonlinear function $\mathbf{g}$.
However, two challenges remain that may prevent their direct adoption in adaptive filters.
First, the result of the constrained maximization is difficult to interpret.
Intuitively, the maximum should give a sort of splitting criterion.
However, the selection of that splitting threshold is not immediately obvious.
Second, the previously described metrics are not invariant under changes of units and rely on nondimensionalizations to make sense if the vectors considered consist of variables that have different units.

In reality, the function $\mathbf{g}$ might map to a space where different elements of the vector have very different scales.
For example, the Cartesian to polar transformation will result in output that has bounded angles and unbounded arbitrarily large distances.
This might promote splitting directions which minimize the approximation error of the final distribution in the radial direction but leave a poor approximation in angle space.
Velocities and positions on differing scales are another example that provides potential difficulties, though an appropriate normalization may solve these issues.
Instead of requiring an application dependent normalization/nondimensionalization, we point out that there is another convenient and natural metric to consider in this problem beyond the Euclidean metric.
In particular, applying the \textit{Mahalanobis distance} carefully to our optimization problem will solve both problems, providing an interpretable splitting criterion while also making the splitting direction choice invariant under rescalings of the coordinates.

The previous discussion on uncertainty based weighting of the optimization was shown to be computationally equivalent to a whitening transformation on the input space.
We now employ an additional whitening transformation on the \textit{output} space as well, using the linearly propagated covariance to induce the new Mahalanobis distance based metric.

Rather than using the standard Euclidean 2-norm on vectors in $\mathbb{R}^m$ from the output of $\mathbf{g}$ or the terms in its Taylor series approximation, we use the norm induced by the linearly predicted precision matrix of the output. The linear prediction of the covariance of $\mathbf{g}(\mathbf{x})$ is
\begin{equation}
    \mathbf{P}_z=\mathbf{G}\mathbf{P}_x\mathbf{G}^T
\end{equation}
and the precision matrix is given by
\begin{equation}
    \label{eq:output_precision_matrix}
    \mathbf{P}_z^{-1}=\mathbf{G}^{-T}\mathbf{P}_x^{-1}\mathbf{G}^{-1}
\end{equation}
if the transformation $\mathbf{G}$ is invertible.
Thus, the norm induced by the final precision matrix is
\begin{equation}
    \Vert\mathbf{z}\Vert_{\mathbf{P}_z^{-1}}^2=\mathbf{z}^T\mathbf{P}_z^{-1}\mathbf{z}
\end{equation}
In the next section, we will demonstrate how this Mahalanobis distance norm applies to each of the splitting methods described so far. We will refer to methodologies for splitting direction selection that employ this output whitening process as Whitened or with an addition of the letter ``W" at the beginning of the acronym.
\subsection{SPHERICAL-AVERAGE-BASED HEURISTICS}
\label{sec:spherical_average}
For a given nonlinear function $\mathbf{g}$, the chosen objective function may admit multiple non-orthogonal maxima.
In such cases, it may be advantageous to choose an alternative direction that aligns well with directions of strong nonlinearity or stretching on average. See Fig. \ref{fig:polar_sos_solc} in Sec. \ref{sec:applications} for more motivation of nearby non-orthogonal peaks of nonlinearity in the example of the Cartesian to polar transformation.
Denoting the base objective function by $\mathcal{H}(\mathbf{g}, \mathbf{x})$, we then consider the original optimization problem \eqref{eq:unit_constrained_opt} but with respect to the spherical-average generalized objective function
\begin{align}
  \label{eq:spherical_average}
    \mathcal{F}(\mathbf{g}, \mathbf{x}) = \int_{\mathbf{u}^T\mathbf{P}_x^{-1} \mathbf{u}=1} (\mathbf{u}^{T}\mathbf{x})^{2} \mathcal{H}^{2}(\mathbf{g},\mathbf{u}) \mathrm{d} \varphi(\mathbf{u})
\end{align}
where $\varphi(\mathbf{u})$ is the ($n-1$)-dimensional surface measure on the ellipsoid $\{\mathbf{u} \, : \, \mathbf{u}^{T} \mathbf{P}_{x}^{-1} \mathbf{u}=1\}$.
The integral form combined with the inner product output scaling in~\eqref{eq:spherical_average} captures the directional nonlinearity or stretching along $\mathbf{x}$ as well as the nonlinearity or stretching along neighboring directions.
\subsubsection{FIRST-ORDER AVERAGING}

Consider the following generalization of the \ac{fos} optimization:
\begin{align}
    \label{eq:safos}
    \hat{\mathbf{x}}^*\sim \argmax_{\|\mathbf{x}\|=1} \int_{\mathbf{u}^T\mathbf{P}_x^{-1} \mathbf{u}=1} (\mathbf{u}^{T}\mathbf{x})^{2}\Vert \mathbf{G}\mathbf{u}\Vert_{2}^{2} \mathrm{d} \varphi(\mathbf{u})
\end{align}
which identifies the direction with the strongest alignment with stretched directions on average over the 1-$\sigma$ uncertainty hyperellipsoid.
Letting $\mathbf{M}=(\mathbf{G}\mathbf{P}_{x}^{1/2})^{T}(\mathbf{G}\mathbf{P}_{x}^{1/2})$, $\mathbf{y}=\mathbf{P}_{x}^{-1/2} \mathbf{u}$, and $\mathbf{a}=\mathbf{P}_{x}^{T/2} \mathbf{x}$, \eqref{eq:safos} can be rewritten as
\begin{align}
    \hat{\mathbf{x}}^*\sim \argmax_{\|\mathbf{x}\|=1} \int_{S_{n}} (\mathbf{a}^{T} \mathbf{y})^{2}\mathbf{y}^{T}\mathbf{M}\mathbf{y}
\mathrm{d} \varphi(\mathbf{y})
\end{align}
where the integral is taken over the origin-centered hypersphere
\begin{align}
 S_n=\left\{\mathbf{y} \in \mathbb{R}^n:\|\mathbf{y}\|=1\right\}
\end{align}
and $\varphi(\mathbf{y})$ is the $(n-1)$-dimensional surface measure on $S_{n}$.
Examining the integral,
\begin{align}
\int_{S_{n}} &\left(\mathbf{a}^T \mathbf{y}\right)^2 \mathbf{y}^T \mathbf{M} \mathbf{y} \mathrm{d} \varphi(\mathbf{y})
=
a_i a_j M_{l,k} \int_{S_{n}} y^i y^j y^k y^l \mathrm{d} \varphi(\mathbf{y})\\
&=\frac{\pi^2}{12} a_i a_j M_{l,k} (\delta^{ij}\delta^{kl} + \delta^{ik}\delta^{jl} + \delta^{il}\delta^{jk})\\
&\propto \mathbf{a}^{T}(\operatorname{tr}(\mathbf{M})\mathbf{I} + 2\mathbf{M})\mathbf{a}
\end{align}
Thus, the optimal split direction is given by
\begin{align}
\hat{\mathbf{x}}^*
&\sim \argmax_{\|\mathbf{x}\|=1} \mathbf{x}^{T}(\operatorname{tr}(\mathbf{M})\mathbf{P}_{x} + 2\mathbf{P}_{x}\mathbf{G}^{T}\mathbf{G}\mathbf{P}_{x})\mathbf{x}
\end{align}
We refer to this method as \ac{safos}.
Note that if $\mathbf{P}_x$ is a scaling of the identity matrix and thus the distribution is isotropic, this method becomes identical to the \ac{fos} method.

Interestingly, the maximizer of the objective function for \ac{safos} in \eqref{eq:spherical_average} when $\mathcal{H}^{2}(\mathbf{g},\mathbf{u})$ is a polynomial in $\mathbf{u}$ is identical to the maximizer for the objective function
\begin{align}
  \label{eq:spherical_average_expectation}
    \mathcal{F}(\mathbf{g}, \mathbf{x}) = \mathrm{E}_{\mathbf{u}}[(\mathbf{u}^{T}\mathbf{x})^{2} \mathcal{H}^{2}(\mathbf{g},\mathbf{u})]
\end{align}
where the second argument denotes that expectation is with respect to the variable $\mathbf{u}$ which is distributed with mean zero and covariance $\mathbf{P}_x$.
This is because, given two polynomials of the same order, the ratio of their expectations with respect to a random multivariate normal variable with covariance $\mathbf{P}$ is equal to the ratio of the surface integrals of the polynomials over the 1-$\sigma$ ellipsoid corresponding to $\mathbf{P}$.
\subsubsection{SECOND-ORDER AVERAGING}
Choosing the functional for averaging to represent the second-order contribution to the nonlinearity
\begin{equation}
    \mathcal{H}^2(\mathbf{g},\mathbf{u})=\Vert\mathbf{G}^{(2)}\mathbf{u}^2\Vert^2_2
\end{equation}
we develop another method for splitting direction selection.
Following a similar procedure to develop \ac{safos}, we propose the \ac{sasos} heuristic
\begin{align}
\hat{\mathbf{x}}^*
&\sim \argmax_{\|\mathbf{x}\|=1} \mathbf{x}^{T}\mathbf{Q}\mathbf{x}
\end{align}
where the quadratic form $\mathbf{Q}$ is given by
\begin{align}
    Q^{i_1,i_2}=(\mathbf{G}^{(2)})^p_{i_3,i_4}(\mathbf{G}^{(2)})^q_{i_5,i_6}\delta_{p,q}C^{i_1...i_6}
\end{align}
and
\begin{align}
    (\mathbf{C}')^{i_1...i_6}&=(\mathbf{P}_x)^{i_1,i_2}(\mathbf{P}_x)^{i_3,i_4}(\mathbf{P}_x)^{i_5,i_6}\\
    \mathbf{C}&=\mathrm{sym}(\mathbf{C}')
\end{align}
where $\mathrm{sym}(\cdot)$ denotes the symmetrization of the given tensor--the average over all permutations of the indices.
In the case where $\mathbf{P}_x$ is isotropic, the objective function can be written in the form
\begin{align}
    x_{i_1}x_{i_2}Q^{i_1,i_2}=\alpha x_{i_1}x_{i_2}\delta^{i_1,i_2}\nonumber\\
    +\beta x_{i_1}x_{i_2}(\mathbf{G}^{(2)})^p_{i_3,i_4}(\mathbf{G}^{(2)})^q_{i_5,i_6}\delta_{p,q}\delta^{i_1,i_3}\delta^{i_2,i_4}\delta^{i_5,i_6}
\end{align}
where $\alpha,\beta$ are constants independent of $\mathbf{x}$ omitted for brevity. The first term is always the constant $\alpha$ for $\mathbf{x}$ on the unit sphere (over which the optimization takes place). The second term can be recognized as being proportional to
\begin{align}
    \mathrm{tr}(\mathbf{D})=\Vert\mathbf{G}^{(2)}\mathbf{x}\Vert_F^2
\end{align}
where $\mathbf{D}$ is defined as
\begin{equation}
    D_{i_5,i_6}=x_{i_1}x_{i_2}\hat{\mathbf{G}}^{(2)}_{i_3,i_4,i_5,i_6}\delta^{i_1,i_3}\delta^{i_2,i_4}
\end{equation}
and
\begin{equation}
    \hat{\mathbf{G}}^{(2)}=\mathrm{sym}(\tilde{\mathbf{G}}^{(2)})
\end{equation}
Thus, in the isotropic case, \ac{sasos} will choose the same direction as \ac{solc} since both optimize the Frobenius norm of the second-order change in the Jacobian in this case.
In that sense, \ac{sasos} can be viewed as a generalization of the \ac{solc} based methods and not a generalization of the \ac{sos} methods.

\subsection{HYBRID METHODOLOGIES}
\label{sec:hybrid_methodologies}
Here we bring together heuristics informed by the nonlinear function properties, initial uncertainty, and weighting from the linear approximation of the final uncertainty.

\subsubsection{LINEAR STRETCHING DIRECTIONS}
To accommodate the best linear stretching direction and weighting based on the initial uncertainty we can construct the following optimization
\begin{equation}
    \hat{\mathbf{x}}^*\sim \argmax_{\mathbf{x}^T\mathbf{P}_x^{-1}\mathbf{x}=1}\Vert \mathbf{G}\mathbf{x}\Vert_2
\end{equation}
which can be solved as the unit vector in the direction of the maximal eigenvector of the symmetric generalized eigenvalue problem associated with the matrices $(\mathbf{G}^T\mathbf{G},\mathbf{P}_x^{-1})$ or as the unit vector in the direction of $\mathbf{P}_x^{1/2}\mathbf{v}$, where $\mathbf{v}$ represents the maximal right singular vector of $\mathbf{G}\mathbf{P}_x^{1/2}$.
We refer to this method as \ac{usfos}.

For this particular heuristic, we do \textit{not} consider the aforementioned output normalization in terms of the linear predicted Mahalanobis distance.
This is because the result is that all directions are equal candidates for splitting and the optimization problem
\begin{equation}
    \hat{\mathbf{x}}^*\sim \argmax_{\mathbf{x}^T\mathbf{P}_x^{-1}\mathbf{x}=1}\Vert \mathbf{G}\mathbf{x}\Vert_{\mathbf{P}_z^{-1}}
\end{equation}
has no unique solution.
Specifically, every element of an initial covariance ellipsoid gets mapped through the linearization onto an associated covariance ellipsoid in the output space that has constant Mahalanobis distance from the output mean.
That is, $\Vert\mathbf{G}\mathbf{x}\Vert_{\mathbf{P}_z^{-1}}$ takes a constant value for the set of $\mathbf{x}$ such that $\mathbf{x}^T\mathbf{P}_x^{-1}\mathbf{x}=1$.
Thus, the specific combination of linearized stretching directions with whitening of the input and the output spaces is meaningless to consider for splitting direction selection.
\subsubsection{SECOND-ORDER STRETCHING DIRECTIONS}
\label{sec:second-order}
The \ac{wussos} heuristic
\begin{equation}
  \label{eq:hybrid_second_order_stretching}
    \hat{\mathbf{x}}^*\sim \argmax_{\mathbf{x}^T\mathbf{P}_x^{-1}\mathbf{x}=1}\Vert \mathbf{G}^{(2)}\mathbf{x}^2\Vert_{\mathbf{P}_z^{-1}}
\end{equation}
 combines nonlinearity, uncertainty, and output whitening to answer the question: What is the maximum Mahalanobis distance (induced by the linearly propagated covariance) of the second-order approximation of the linearization error for any initial point on a surface of equal likelihood in the input space?
The \ac{wussos} optimization problem can be rephrased in a manner computable as a maximal Z-eigenvector with shifted symmetric higher-order power iteration as
\begin{equation}
    \hat{\mathbf{x}}^*\sim \mathbf{P}_x^{1/2}\argmax_{\mathbf{y}^T\mathbf{y}=1}  (\mathbf{G}^{(2)}(\mathbf{P}_x^{1/2}\mathbf{y})^2)^T \mathbf{P}_z^{-1} (\mathbf{G}^{(2)}(\mathbf{P}_x^{1/2}\mathbf{y})^2)
    \label{eq:whitened_second_order}
\end{equation}
\subsubsection{CHANGE IN LINEARIZATION}
The original work by Tuggle and Zanetti \cite{tuggle2018automated,tuggle2020model} employed the first two facets of nonlinear function properties and uncertainty scaling to arrive at the optimization
\begin{equation}
    \hat{\mathbf{x}}^*\sim \argmax_{\mathbf{x}^T\mathbf{P}_x^{-1}\mathbf{x}=1}\Vert \mathbf{G}^{(2)}\mathbf{x}\Vert_{F}
\end{equation}
which can be solved in terms of a generalized eigenvalue problem with the two matrices $(\bar{\mathbf{G}}^{(2)},\mathbf{P}_x^{-1})$ or as the unit vector in the direction of $\mathbf{P}_x^{1/2}$ multiplied with the maximal right singular vector of the matrix $\bar{\mathbf{G}}^{(2)}\mathbf{P}_x^{1/2}$.

We can establish a scale invariance for this selection criterion by transforming the linear transformation given by the matrix $\mathbf{G}^{(2)}\mathbf{x}$ so that it maps from a whitened input space to a whitened output space.
The input whitening transformation is given by
\begin{equation}
    \mathbf{x}'=\mathbf{P}_x^{-1/2}\mathbf{x}
\end{equation}
and the whitening transformation for the output space is given by
\begin{equation}
    \mathbf{z}'=\mathbf{P}_z^{-1/2}\mathbf{z}
\end{equation}
A generic linear transformation $\mathbf{A}$ from the original input space to the original output space can be transformed to a linear transformation from the whitened input space to the whitened output space according to
\begin{equation}
    \mathbf{A}'= \mathbf{P}_z^{-1/2} \mathbf{A} \mathbf{P}_x^{1/2}
\end{equation}
If we apply this transformation to the linear transformation given by the matrix $\mathbf{G}^{(2)}\mathbf{x}$ we obtain the matrix
\begin{equation}
    \mathbf{P}_z^{-1/2}(\mathbf{G}^{(2)}\mathbf{x})\mathbf{P}_x^{1/2}
\end{equation}
whose squared Frobenius norm will characterize the change in the linear approximation of the nonlinear function $\mathbf{g}$ (normalized to map between whitened input and output spaces) at a step $\mathbf{x}$ away from the current point of linearization. In order to assess whether the change in this whitened linearization is significant we should compare against the squared Frobenius norm of the original linearization between whitened spaces.

If the original linear transformation is $\mathbf{A}=\mathbf{G}$ and both $\mathbf{P}_x,\mathbf{P}_z$ are nonsingular, then the corresponding linear transformation between whitened spaces $\mathbf{G}'$ is a linear transformation from the unit sphere in $n$ dimensions to the unit sphere in $m$ dimensions.
Thus, $\mathbf{G}'$ is an orthogonal matrix. Note that since the eigenvalues of an orthogonal matrix only take values $-1,0$, and $1$, the rank of the matrix gives the number of nonzero eiegnvalues, and the Frobenius norm of $\mathbf{G}'$ is given by the sum of squares of the eigenvalues
\begin{equation}
\Vert\mathbf{G}'\Vert_F^2=\min(n,m)
\end{equation}
As such, the squared Frobenius norm of the whitened change in the linearization given by $\mathbf{G}^{(2)}\mathbf{x}$ can naturally be compared to $\min(n,m)$. If this value is small relative to $\min(n,m)$, then the linearization at a point $\mathbf{x}$ away from the mean would propagate the initial covariance ellipsoid to a final covariance ellipsoid that is very similar.

To select a direction in which the change in linearization will affect the propagation of covariance to the highest degree, we propose the \ac{wussolc} method, which performs the optimization
\begin{equation}
    \hat{\mathbf{x}}^*\sim \mathbf{P}_x^{1/2}\argmax_{\mathbf{x}^T\mathbf{P}_x^{-1}\mathbf{x}=1}\Vert \mathbf{P}_z^{-1/2}(\mathbf{G}^{(2)}\mathbf{x})\mathbf{P}_x^{1/2}\Vert_{F}^2
\end{equation}
By defining the related tensor (whose ``square'' is the tensor at play in~\eqref{eq:whitened_second_order})
\begin{equation}
    (\mathbf{G}_w^{(2)})^i_{j,k}=(\mathbf{P}_z^{-1/2})^i_l(\mathbf{G}^{(2)})^l_{p,q}(\mathbf{P}_x^{1/2})^p_j(\mathbf{P}_x^{1/2})^q_k
\end{equation}
and its $mn$ by $n$ matricization
\begin{equation}
    (\bar{\mathbf{G}}^{(2)}_w)^{ni+j}_k=(\mathbf{G}^{(2)}_w)^i_{j,k}
\end{equation}
we can write the splitting direction as
\begin{equation}
    \hat{\mathbf{x}}^*\sim \mathbf{P}_x^{1/2}\argmax_{\mathbf{y}^T\mathbf{y}=1}\Vert \bar{\mathbf{G}}^{(2)}_w\mathbf{y}\Vert_{2}^2
\end{equation}
With this, the splitting direction is obtained as the unit vector in the direction of $\mathbf{P}_x^{1/2}\mathbf{v}$, where $\mathbf{v}$ represents the maximal right singular vector of $\bar{\mathbf{G}}^{(2)}_w$. An effective splitting criterion might test the value
\begin{equation}
    \frac{1}{\min(n,m)}\max_{\mathbf{x}^T\mathbf{P}_x^{-1}\mathbf{x}=1}\Vert \mathbf{P}_z^{-1/2}(\mathbf{G}^{(2)}\mathbf{x})\mathbf{P}_x^{1/2}\Vert_{F}^2
\end{equation}
as a sort of maximal normalized change in the linearization.

\subsubsection{STATISTICAL AND DETERMINISTIC LINEARIZATION DIFFERENCING}
We can extend the method based on differencing the statistical and deterministic linearization to factor in initial uncertainty and rescale coordinates using the whitening method described earlier.
The resulting splitting direction is then
\begin{equation}
    \hat{\mathbf{x}}^*\sim \mathbf{P}_x^{1/2}\argmax_{\mathbf{y}^T\mathbf{y}=1}\Vert \mathbf{P}_z^{-1/2}(\mathbf{G}^{(SL)}-\mathbf{G})\mathbf{P}^{1/2}_x\mathbf{y}\Vert_{2}
    \label{eq:stat_lin_dif_full}
\end{equation}
which can be computed by normalizing the product of $\mathbf{P}_x^{1/2}$ with the maximal right singular vector of the matrix that is a part of the objective in~\eqref{eq:stat_lin_dif_full}.
The covariance $\mathbf{P}_z$ above can either be the covariance obtained from the sigma point method or the covariance given by propagating the original covariance using the deterministic linearization. In this work, we will employ the covariance arising from the deterministic linearization as it arises from assuming Gaussianity of $\mathbf{z}$, and the whitening with a Gaussian distribution better represents a meaningful Mahalanobis distance. This direction is the one in which an input on an $\mathbf{x}$-covariance ellipsoid produces the largest error in the statistical and deterministic linearizations in terms of its $\mathbf{P}_z$-Mahalanobis distance.
This method will be referred to as the \ac{wussadl} method.

\subsubsection{SPHERICAL AVERAGE SECOND-ORDER STRETCHING}
As a whitened extension of \ac{sasos}, we present \ac{wsasos} which is only different from \ac{sasos} in the quadratic form being employed for optimization. Again,
\begin{align}
\hat{\mathbf{x}}^*
&\sim \argmax_{\|\mathbf{x}\|=1} \mathbf{x}^{T}\mathbf{Q}_w\mathbf{x}
\end{align}
where the new quadratic form $\mathbf{Q}_w$ is given by
\begin{align}
    Q^{i_1,i_2}_w=(\mathbf{G}^{(2)})^p_{i_3,i_4}(\mathbf{G}^{(2)})^q_{i_5,i_6}(\mathbf{P}_z^{-1})_{p,q}C^{i_1...i_6}
\end{align}
with the linearly predicted inverse covariance matrix employed in place of the standard Euclidean metric tensor used in \ac{sasos}.

A summary of the splitting heuristics discussed in the paper is provided in Table~\ref{tab:summary_splitting_heuristics}.
\begin{table}[htbp]
    \centering
    \caption{Summary of splitting heuristics.}
    \label{tab:summary_splitting_heuristics}
      \setlength{\tabcolsep}{2pt}
    \begin{tabular}{ccc}
    \toprule[2pt]%
         Heuristic & Objective, $\mathcal{F}(\mathbf{g},\mathbf{x})$  & Constraint \\
      \midrule
         maxvar & $\Vert \mathbf{x} \Vert$ & $\Vert \mathbf{x} \Vert_{\mathbf{P}_{x}^{-1}}$ \\
      \acs{fos}   & $\left\Vert\mathbf{G}\mathbf{x}\right\Vert_2$ & $\|\mathbf{x}\|_2$\\
      \acs{sos}   & $\Vert\mathbf{G}^{(2)}\mathbf{x}^2\Vert_2$ & $\|\mathbf{x}\|_2$\\
      \acs{solc} \cite{tuggle2018automated,tuggle2020model} & $\Vert\mathbf{G}^{(2)}\mathbf{x}\Vert_F$ & $\|\mathbf{x}\|_2$\\
      \acs{sadl}  & $\Vert(\mathbf{G}^{(\textrm{SL})}-\mathbf{G})\mathbf{x}\Vert_2$ & $\|\mathbf{x}\|_2$\\
      \acs{usfos} & $\left\Vert\mathbf{G}\mathbf{x}\right\Vert_2$ & $\Vert\mathbf{x}\Vert_{\mathbf{P}^{-1}_x}$\\
      \acs{ussolc} \cite{tuggle2018automated,tuggle2020model}& $\Vert\mathbf{G}^{(2)}\mathbf{x}\Vert_F$ & $\Vert\mathbf{x}\Vert_{\mathbf{P}^{-1}_x}$\\
      \acs{safos} & $\int\limits_{\mathbf{u}^T\mathbf{P}_x^{-1} \mathbf{u}}\!\!\!\!\!\! (\mathbf{u}^{T}\mathbf{x})^{2}\Vert \mathbf{G}\mathbf{u}\Vert_{2}^{2} \mathrm{d} \varphi(\mathbf{u})$ & $\|\mathbf{x}\|_2$
\\
      \acs{sasos} & $\int\limits_{\mathbf{u}^T\mathbf{P}_x^{-1} \mathbf{u}}\!\!\!\!\!\! (\mathbf{u}^{T}\mathbf{x})^{2}\Vert \mathbf{G}^{(2)}\mathbf{u}^2\Vert_{2}^{2} \mathrm{d} \varphi(\mathbf{u})$ & $\|\mathbf{x}\|_2$\\
      \acs{wussos} & $\Vert \mathbf{G}^{(2)}\mathbf{x}^2\Vert_{\mathbf{P}_z^{-1}}$ & $\Vert\mathbf{x}\Vert_{\mathbf{P}^{-1}_x}$\\
      \acs{wussolc} & $\Vert \mathbf{P}_z^{-1/2}(\mathbf{G}^{(2)}\mathbf{x})\mathbf{P}_x^{1/2}\Vert_{F}^2$ & $\Vert\mathbf{x}\Vert_{\mathbf{P}^{-1}_x}$\\
      \acs{wussadl} & $\Vert \mathbf{P}_z^{-1/2}(\mathbf{G}^{(\textrm{SL})}-\mathbf{G})\mathbf{x}\Vert_{2}$ & $\Vert\mathbf{x}\Vert_{\mathbf{P}^{-1}_x}$\\
      \acs{wsasos} & $\int\limits_{\mathbf{u}^T\mathbf{P}_x^{-1} \mathbf{u}}\!\!\!\!\!\! (\mathbf{u}^{T}\mathbf{x})^{2}\Vert \mathbf{G}^{(2)}\mathbf{u}^2\Vert_{\mathbf{P}_x^{-1}}^{2} \mathrm{d} \varphi(\mathbf{u})$ & $\|\mathbf{x}\|_2$\\
    \bottomrule[2pt]
    \end{tabular}
\end{table}

\subsection{Recursive Splitting}

\Ac{gm} splitting is achieved through application of the operator $\tilde{G}_{c}$, which operates on an arbitrary multivariate \ac{gm} $p(\mathbf{x})$ as
\begin{align}
  \tilde{G}_{c}[p(\mathbf{x})] = \sum_{i = 1}^{I} G_{c}[w_{i}\mathcal{N}(\mathbf{x};\boldsymbol{\mu}_{i},\mathbf{P}_{i})]
\end{align}
where
\begin{align}
  G_{c}&[w_{i}\mathcal{N}(\mathbf{x};\boldsymbol{\mu}^{(i)},\mathbf{P}_{i})]\nonumber \\
       &= \begin{cases}
        w_{i}\mathcal{N}(\mathbf{x};\boldsymbol{\mu}_{i},\mathbf{P}_{i}) & c(w_{i},\boldsymbol{\mu}_{i},\mathbf{P}_{i}) \leq 0\\
        \sum\limits_{\ell=1}^{L}G_{c}[w_{i,\ell}\mathcal{N}(\mathbf{x};\boldsymbol{\mu}_{i,\ell},\mathbf{P}_{i,\ell})] & \mathrm{otherwise}
    \end{cases}
\end{align}
and $c(\omega^{(i)},\boldsymbol{\mu}^{(i)},\mathbf{P}^{(i)})$ is a splitting criterion  \cite{iannamorelli2024AdaptiveGaussianMixture}.
The operator is applied recursively on all newly generated mixands until no mixands satisfy the splitting criterion.
In the simplest case, the criterion function $c$ corresponds directly to the split direction heuristic maximal value.
For example, the splitting criterion for the \ac{wussos} heuristic~\eqref{eq:hybrid_second_order_stretching} is
\begin{align}
  c(w,\boldsymbol{\mu},\mathbf{P}_{x})
  =
  \| \mathbf{G}^{(2)} (\hat{\mathbf{x}}^{*})^{2} \|_{\mathbf{P}_{z}^{-1}}
\end{align}
where $\hat{\mathbf{x}}^{*}$ is the maximizing direction and $\mathbf{P}_{z}^{-1}$ is given in~\eqref{eq:output_precision_matrix} with the Jacobian $\mathbf{G}$ and second-order partial derivative tensor $\mathbf{G}^{(2)}$ evaluated at the mixand mean $\boldsymbol{\mu}$.
This criterion does not consider the mixand weight, and consequently may split low weight mixands with negligible contribution to the overall distribution.
Instead, inspired by  \cite{huber2011AdaptiveGaussianMixture}, the geometric average
\begin{align}
  c(w,\boldsymbol{\mu},\mathbf{P}_{x})
  =
  w^{\gamma} \| \mathbf{G}^{(2)} (\hat{\mathbf{x}}^{*})^{2} \|_{\mathbf{P}_{z}^{-1}}^{(1-\gamma)}
\end{align}
is employed with $\gamma \in [0,1]$, which jointly considers the mixand weight and degree of nonlinearity.
When $\gamma=0$, only the original heuristic is considered, whereas only the mixand weight is considered when $\gamma=1$.

\section{METRICS FOR EVALUATION}
In order to assess the methods presented in this paper for uncertainty propagation, we compare against Monte Carlo results or analytical solutions for the propagated distribution as baseline ``truth.''
If an exact analytic solution of the output \ac{pdf} is available, the \ac{gm} approximation error is directly measured via the \ac{nise}
\begin{align}
  \texttt{NISE} = \frac{\int (p(\mathbf{z}) - p'(\mathbf{z}))^{2} \mathrm{d} \mathbf{z}}{\int p^{2}(\mathbf{z}) \mathrm{d} \mathbf{z} + \int p'^{2}(\mathbf{z}) \mathrm{d} \mathbf{z}}
\end{align}
where $p,p'$ are \acp{pdf} for the two random variables being compared \cite{williams2003GaussianMixtureReduction}.

In the case where the true distribution is not known analytically and only $N$ samples $\mathbf{z}_{(i)}'$ from the ``truth'' distribution of the random variable $\mathbf{z}'$ can be obtained from Monte Carlo analysis, we employ a number of other metrics.
For univariate distributions, the \ac{cvm} metric provides a measure of the goodness of fit between a given distribution and an empirical approximation \cite{darling1957KolmogorovSmirnovCramervonMises}.
The \ac{cvm} metric between the cumulative marginal distribution $F^{*}(z^{j})$ and $N$-sample empirical distribution $F_{N}(z^{j})$ is given by
\begin{align}
    \omega_{j}^2 = \int_{-\infty}^{\infty} \left(F^{*}(z^{j}) - F_{N}(z^{j})\right)^{2} \mathrm{d} F^{*} (z^{j})
\end{align}
For multivariate distributions, we measure the overall goodness-of-fit by the 2-norm of the marginal \ac{cvm} metrics arranged as a vector:
\begin{align}
  \|\boldsymbol{\omega}^{2}\|_{2} = \left\Vert
  \begin{bmatrix}
    \omega_{1}^{2} & \cdots & \omega_{m}^{2}
  \end{bmatrix}
  \right\Vert_{2}
\end{align}

The errors in the means and covariances respectively constitute alternative measures of performance.
We introduce the \ac{MaDEM}
\begin{equation}
    \Vert\boldsymbol{\mu}_z-\boldsymbol{\mu}_z'\Vert_{\mathbf{P}_z^{-1}}
\end{equation}
which serves as as an interpretable metric.
For reference, the \ac{MaDEM} can be compared to the dimension of $\mathbf{z}$ to determine if the mean shifted appreciably relative to the linearly predicted uncertainty.

\begin{figure}
    \centering
\includegraphics[width=0.5\linewidth]{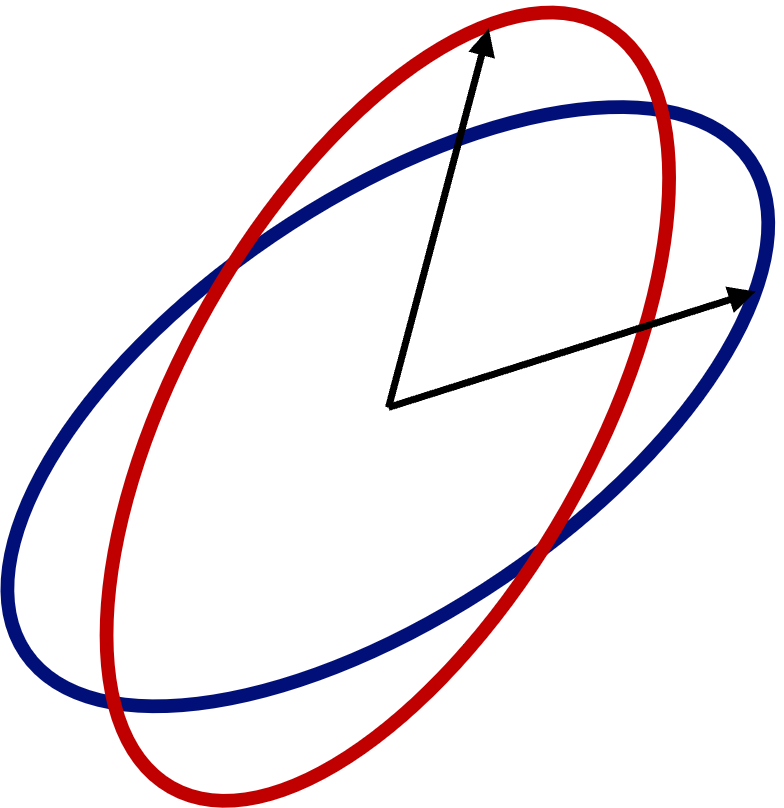}
    \caption{Notional depiction of the directions along which minimum and maximum ratios of two ellipses occur}
    \label{fig:ellipse-ratio}
\end{figure}

To understand the error of the covariance, we also introduce a measure of the level of disagreement of the 1-sigma covariance ellipsoids.
This disagreement is best quantified by the maximum/minimum ratio along any direction of the ellipsoids given by the two equations
\begin{align}
    \mathbf{z}^T\mathbf{P}_z^{-1}\mathbf{z}&=1\\
    \mathbf{z}^T(\mathbf{P}_z')^{-1}\mathbf{z}&=1
\end{align}
A notional depiction of the maximum and minimum ratios of two ellipses is depicted along two vectors in Fig. \ref{fig:ellipse-ratio}.
Suppose that $\rho\hat{\mathbf{z}}$ is the vector along the 1-sigma contour associated with $\mathbf{P}_z$, where $\hat{\cdot}$ denotes a unit vector. Similarly, suppose that $\rho'\hat{\mathbf{z}}$ is the vector along the 1-sigma contour associated with $\mathbf{P}_z'$. Then
\begin{align}
    \frac{\rho^2\hat{\mathbf{z}}^T\mathbf{P}_z^{-1}\hat{\mathbf{z}}}{(\rho')^2 \hat{\mathbf{z}}^T(\mathbf{P}_z')^{-1}\hat{\mathbf{z}}}&=1\implies\\
    \frac{\hat{\mathbf{z}}^T\mathbf{P}_z^{-1}\hat{\mathbf{z}}}{ \hat{\mathbf{z}}^T(\mathbf{P}_z')^{-1}\hat{\mathbf{z}}}&=\frac{(\rho')^2}{\rho^2}
    \label{eq:gen_rayleigh}
\end{align}
Thus, the generalized Rayleigh quotient in~\eqref{eq:gen_rayleigh} gives the squared ratio of the magnitude of the covariance ellipsoids along the direction $\hat{\mathbf{z}}$. This generalized Rayleigh quotient is extremized at the generalized eigenvectors of the generalized eigenproblem
\begin{equation}
    \mathbf{P}^{-1}_z\mathbf{z}=\lambda(\mathbf{P}_z')^{-1}\mathbf{z}
    \label{eq:gen_eig}
\end{equation}
The corresponding eigenvalues give the stationary values of the ratios of the two ellipsoids along any directions.
This can be obtained by applying the method of Lagrange multipliers to the optimization of the Rayleigh quotient over the unit sphere.
The value $\max(1/\min_i\lambda_i,\max_i\lambda_i)$ is then a measure of the error between the two covariances that we will denote as \ac{mcr}.

An \ac{mcr} value of unity indicates that the two covariance are identical, whereas a value above unity indicates that the 1-sigma covariance ellipsoids have that ratio along some direction.
The generalized eigenvalues in~\eqref{eq:gen_eig} can also be computed in two other potentially more convenient fashions as the squares of the singular values of $\mathbf{P}_z^{-1/2}(\mathbf{P}_z')^{1/2}$.
This way, the inverse of both covariances need not be calculated after both Cholesky factors are calculated.
Rather, the result is obtained simply from applying forward substitution to each column of $(\mathbf{P}_z')^{1/2}$.
Or, as the name suggests, the change of variables $\boldsymbol{\zeta}=\mathbf{P}_z\mathbf{z}$ and a multiplication by $\mathbf{P}_z'$ takes~\eqref{eq:gen_eig} to the symmetric generalized eigenvalue problem in terms of the covariances rather than the precisions
\begin{equation}
    \mathbf{P}'_z\boldsymbol{\zeta}=\lambda\mathbf{P}_z\boldsymbol{\zeta}
\end{equation}

The \ac{elk} \cite{jebara2004ProbabilityProductKernels} is another measure of performance employed to compare against an empirical ``truth" distribution.
If $\mathbf{z}$ has probability density $p$ and $\mathbf{z}'$ has probability density $p$ then the \ac{elk} is
\begin{equation}
    l^{2}(p,p')
    =
    \mathrm{E}_{p'}[p(\mathbf{z})]
    =
    \int_\Omega p(\mathbf{z})p'(\mathbf{z})\mathrm{d}\mathbf{z}\approx \frac{1}{N}\sum_{i=1}^N p(\mathbf{z}'_{(i)})
\end{equation}
where $\Omega$ is the support of the probability density functions.
The \ac{elk} is also known as the \ac{lam} in other work (see, e.g., \cite{demars2013entropy}).

\section{APPLICATIONS}
\label{sec:applications}
This section demonstrates the application of the proposed splitting methodology to three distinct nonlinear uncertainty mapping problems.
In each of the examples, a Gaussian input density is split according to each of the aforementioned split methods, after which the resulting \ac{gm} is mapped to the output space via linearization.
As a baseline, we additionally consider two other existing methods.
The first simple approach involves splitting along the principal direction of highest variance and abbreviate this method as ``maxvar.''
The second method, coined the \ac{alodt} by the original authors, measures nonlinearity by the deviation of propagated sigma points from their linear regression fit and splits along the principal direction over which this measure is maximal \cite{faubel2010FurtherImprovementAdaptive}.
To enable direct comparison between splitting heuristics, the mixand selection criteria are ignored, and, instead, all mixands are split to the maximum recursion depth.
By this approach, the output density for each method has an equal number of mixands.

For the \ac{sadl} and \ac{wussadl} methods, we obtain the statistical linearization matrix $\mathbf{G}^{(\textrm{SL})}$ via~\eqref{eq:statistical_linearization_matrix} and the \ac{sut} \cite{julier2002ScaledUnscentedTransformation}.
Because the input distribution is Gaussian, the \ac{sut} scaling parameter value $\beta_{\textrm{SUT}}=2$ is used for optimality \cite{julier2002ScaledUnscentedTransformation} and the relatively large $\alpha_{\textrm{SUT}}=0.5$ is chosen to place the sigma points farther from the mean, accentuating the differences in statistical and deterministic linearization.

\subsection{POLAR TRANSFORMATION}
As a simple example, the polar transformation
\begin{equation}
    [r\quad \theta]^T=\mathbf{g}([x\quad y]^T)=\left[\sqrt{x^2+y^2}\quad \arctan\left(\frac{y}{x}\right)\right]^T
\end{equation}
is employed.
We begin with a Gaussian random variable in Cartesian space with mean $\boldsymbol{\mu}_x=[0 \quad 1000]^T$ and covariance $\mathbf{P}_x=250^2\operatorname{diag}([16 \quad 1])$, where the $\operatorname{diag}(\cdot)$ operator returns a diagonal square matrix with its argument on the diagonal.
These distribution parameters are chosen as a stressing case, in that the distribution of the random variable in polar coordinates is highly non-Gaussian, and such that the resulting distribution in polar coordinates has support on very different scales for the magnitude and angle coordinates.
In this example, we limit the nested splitting to a recursion depth of two, such that the split distribution is composed of nine mixands.
The approximation accuracy of each method, as measured by the \ac{nise}, is tabulated in Table~\ref{tab:cart2polar_results}.

\label{sec:applications-measure}

\begin{table}[htbp]
  \caption{Cartesian to polar coordinate example.}
  \label{tab:cart2polar_results}
  \centering
    \begin{tabular}{lr}
      \toprule[2pt]%
      Method & NISE$\downarrow$\\
      \midrule
      \begin{filecontents*}{cart2polar_results.csv}
method   , ISE                    , NISE   
variance , 2.54e-05               , 0.0354 
FOS      , 0.0001572              , 0.2260 
SOS      , 2.60e-05               , 0.0365 
SOLC     , 3.11e-05               , 0.0441 
SADL     , 2.5539816056163766e-05 , 0.0356 
USFOS    , 0.00015723753434898865 , 0.2260 
USSOLC   , 2.51e-05               , 0.0351 
SAFOS    , 2.63e-05               , 0.0367 
SASOS    , 2.51e-05               , 0.0350 
WUSSOS   , 2.52e-05               , 0.0351 
WUSSOLC  , 2.52e-05               , 0.0351 
WUSSADL  , 2.54e-05               , 0.0354 
WSASOS   , 2.52e-05               , 0.0352 
\end{filecontents*}
\csvreader[head to column names]{cart2polar_results.csv}{}%
      {\method & \NISE\\}\\[-1em]
      \bottomrule
    \end{tabular}
\end{table}

\begin{figure*}
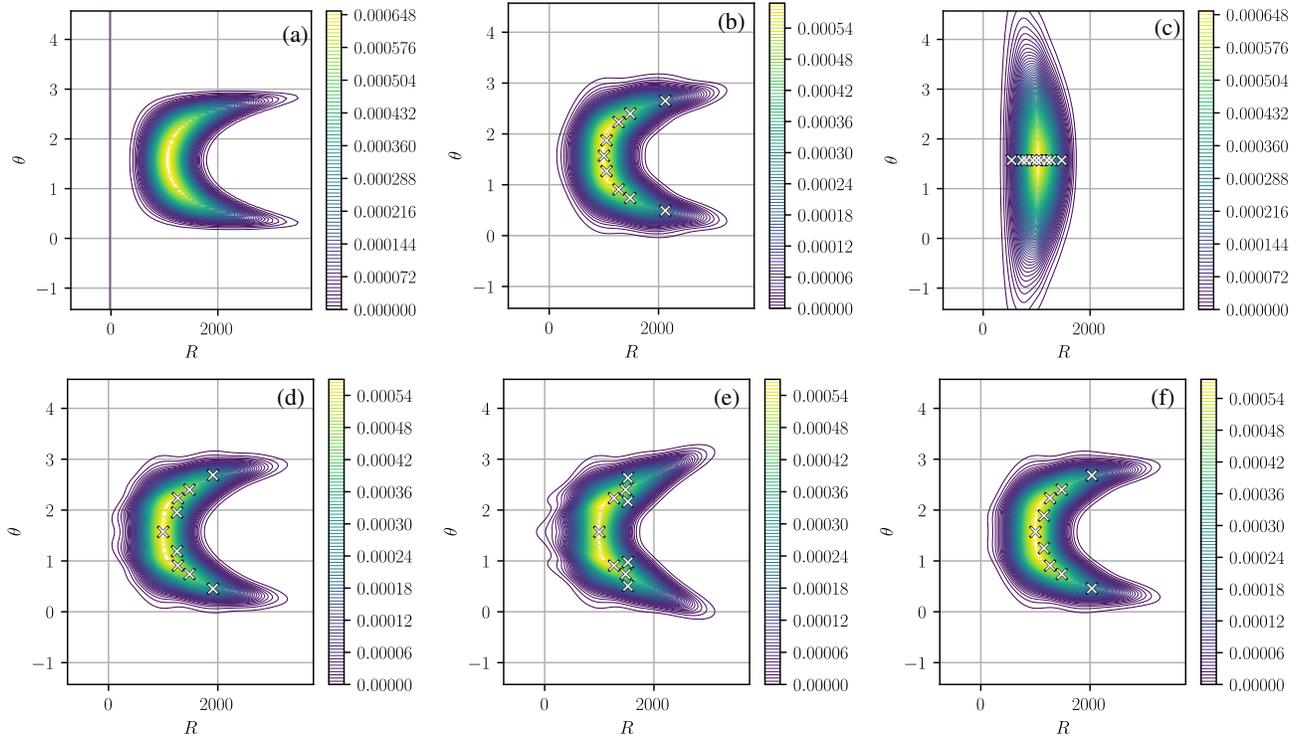

\newcommand{\figwidthfactor}{0.3}
\newcommand{\example}{cart2polar}
\newcommand{\plotsuffix}{_0_1}
\newcommand{\labelxcoord}{-1.8}
\centering
\begin{subcaptiongroup}
  \subcaptionlistentry{throwaway}
  \label{fig:\example:truth}
  \begin{tikzpicture}
    \node[anchor=north east] (img) at (0,0)
    {\includesvg[width=\figwidthfactor\linewidth]{graphics/\example truth.svg}};
    \node at (\labelxcoord,-0.5) {\captiontext*{}};
  \end{tikzpicture}%
  \subcaptionlistentry{throwaway}
  \label{fig:\example:variance}
  \begin{tikzpicture}
    \node[anchor=north east] (img) at (0,0)
    {\includesvg[width=\figwidthfactor\linewidth]{graphics/\example variance\plotsuffix .svg}};
    \node at (\labelxcoord,-0.4) {\captiontext*{}};
  \end{tikzpicture}
  \subcaptionlistentry{throwaway}
  \label{fig:\example:fos}
  \begin{tikzpicture}
    \node[anchor=north east] (img) at (0,0)
    {\includesvg[width=\figwidthfactor\linewidth]{graphics/\example FOS\plotsuffix .svg}};
    \node at (\labelxcoord,-0.4) {\captiontext*{}};
  \end{tikzpicture}
  \subcaptionlistentry{throwaway}
  \label{fig:\example:sos}
  \begin{tikzpicture}
    \node[anchor=north east] (img) at (0,0)
    {\includesvg[width=\figwidthfactor\linewidth]{graphics/\example SOS\plotsuffix .svg}};
    \node at (\labelxcoord,-0.4) {\captiontext*{}};
  \end{tikzpicture}
  \subcaptionlistentry{throwaway}
  \label{fig:\example:solc}
  \begin{tikzpicture}
    \node[anchor=north east] (img) at (0,0)
    {\includesvg[width=\figwidthfactor\linewidth]{graphics/\example SOLC\plotsuffix.svg}};
    \node at (\labelxcoord,-0.4) {\captiontext*{}};
  \end{tikzpicture}
  \subcaptionlistentry{throwaway}
  \label{fig:\example:ussolc}
  \begin{tikzpicture}
    \node[anchor=north east] (img) at (0,0)
    {\includesvg[width=\figwidthfactor\linewidth]{graphics/\example USSOLC\plotsuffix .svg}};
    \node at (\labelxcoord,-0.4) {\captiontext*{}};
  \end{tikzpicture}
\end{subcaptiongroup}
\captionsetup{subrefformat=parens}

\caption{Polar transformation nonlinear-mapped \subref{fig:\example:truth} truth and nonlinear-mapped split densities using \subref{fig:\example:variance} max variance (identical to \ac{alodt}, \ac{sadl}, \ac{safos}, \ac{wsasos}, \ac{wussos}, \ac{wussolc}, \ac{wussadl}), \subref{fig:\example:fos} \ac{fos} (identical to \ac{usfos}), \subref{fig:\example:sos} \ac{sos}, \subref{fig:\example:solc} \ac{solc}, and \subref{fig:\example:ussolc} \ac{ussolc} (identical to \ac{sasos}). Black and white markers indicate mixand mean locations.}
\label{fig:cart2pol_split_transformed}
\end{figure*}
In this situation where the two variable $r,\theta$ have very different scales, the splitting direction selection methods that incorporate output whitening tend to perform roughly the same or slightly better as in the case of \ac{wussos}, \ac{wussolc}, and \ac{wussadl} vs \ac{sos}, \ac{solc}, and \ac{sadl} respectively.
More prominently, \ac{fos} and \ac{usfos} which perform identically, result in the highest mean squared error of all of the methods because they always choose to split along the radial direction.
This supports the idea that first-order splitting methods are not the right choice for all types of nonlinear functions.
The higher-order splitting schemes \ac{wussadl}, \ac{wussolc}, \ac{wussos}, \ac{sasos}, and  \ac{wsasos} perform increasingly better as compared to \ac{fos}.
Comparing the sigma point based methods, the \ac{alodt} method performs similarly but slightly worse than \ac{wussadl}.
Even variance only splitting performs only marginally worse than all but the \ac{fos} based methods for splitting direction selection.

\begin{figure}
    \centering
    \includesvg[width=\linewidth]{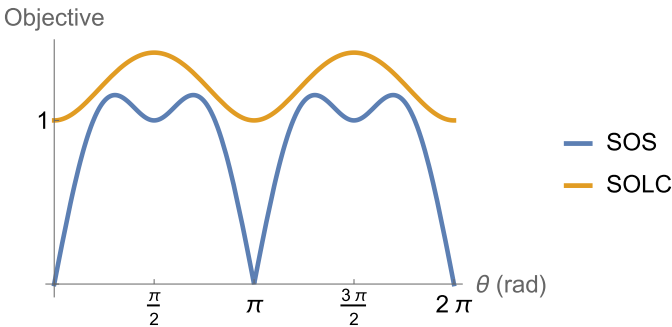}
    \caption{\ac{sos} and \ac{solc} objectives as a function of direction in a simplified polar example.}
    \label{fig:polar_sos_solc}
\end{figure}

Interestingly, the \ac{solc} based methods perform differently in this example than \ac{sos} based methods.
To examine this more closely, the objective function for \ac{sos} and \ac{solc} is shown in Fig.~\ref{fig:polar_sos_solc} as a function of the polar angle of a potential splitting direction.
Specifically, given a point at $[x,y]^T=[1,0]^T$, the objective functions for \ac{sos} and \ac{solc} evaluated for splitting direction $\hat{\mathbf{x}}=[\cos\theta,\sin\theta]^T$, Fig.~\ref{fig:polar_sos_solc} shows the values of the \ac{sos} and \ac{solc} objectives.
We can see that \ac{sos} has two points of equal maximal objective value that are non-orthogonal.
Furthermore, these points are close to, and on either side of, the maximum of the \ac{solc} curve.
The \ac{solc} curve peaks at in the direction orthogonal to the line of sight between the origin and the point $[x,y]^T=[1,0]^T$.
This helps to demonstrate how \ac{sos} and \ac{solc} will select different directions even though they are both based around similar measures of the second-order nonlinearity.
Fig.~\ref{fig:polar_sos_solc}  also highlights the potential nonorthogonality of equivalent maxima of the \ac{sos} objective function.
This non-orthogonality provides additional motivation for the development of \ac{sasos}, which averages out nearby nonorthogonal maxima of \ac{sos} and chooses something in between just as \ac{solc} will.
This effect further supports the notion that \ac{sasos} is a generalization of \ac{solc}.

\subsection{TWO-BODY MOTION}
\begin{figure*}
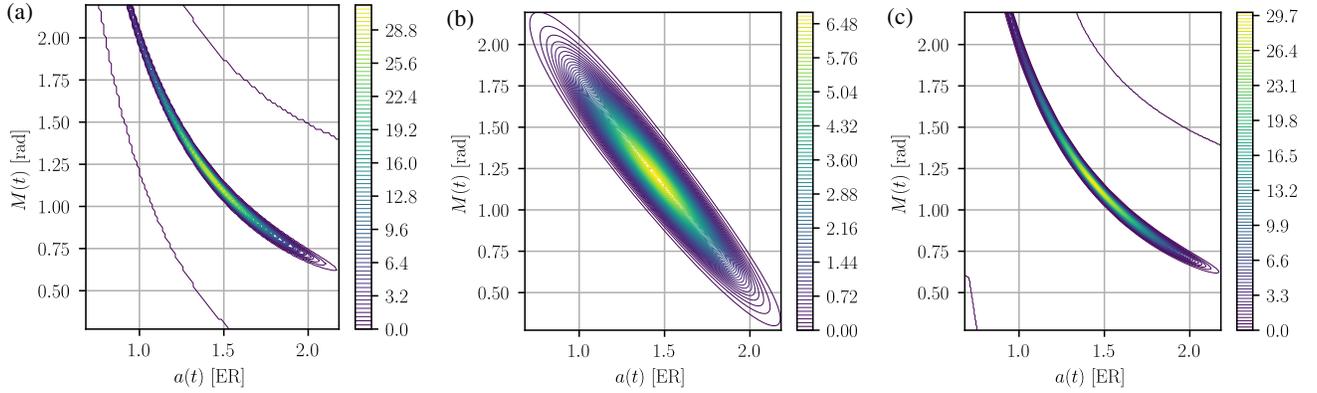

\newcommand{\figwidthfactor}{0.3}
\newcommand{\example}{twobody}
\newcommand{\plotsuffix}{_0_1}
\centering
\begin{subcaptiongroup}
  \subcaptionlistentry{throwaway}
  \label{fig:\example:truth}
  \begin{tikzpicture}
    \node[anchor=north west] (img) at (0,0)
    {\includesvg[width=\figwidthfactor\linewidth]{graphics/\example truth.svg}};
    \node at (0.25,-0.25) {\captiontext*{}};
  \end{tikzpicture}
  \subcaptionlistentry{throwaway}
  \label{fig:\example:fos}
  \begin{tikzpicture}
    \node[anchor=north west] (img) at (0,0)
    {\includesvg[width=\figwidthfactor\linewidth]{graphics/\example FOS\plotsuffix.svg}};
    \node at (0.25,-0.25) {\captiontext*{}};
  \end{tikzpicture}
  \subcaptionlistentry{throwaway}
  \label{fig:\example:usfos}
  \begin{tikzpicture}
    \node[anchor=north west] (img) at (0,0)
    {\includesvg[width=\figwidthfactor\linewidth]{graphics/\example USFOS\plotsuffix.svg}};
    \node at (0.25,-0.25) {\captiontext*{}};
  \end{tikzpicture}
\end{subcaptiongroup}
\captionsetup{subrefformat=parens}
\caption{Keplerian orbital elements nonlinear-mapped \subref{fig:twobody:truth} truth and nonlinear-mapped split densities using \subref{fig:twobody:fos} \ac{fos} and \subref{fig:twobody:usfos} \ac{usfos}. All other plots are identical to the \ac{usfos} plot to the eye.}
\label{fig:two_body}
\end{figure*}

As another example where the nonlinear function $\mathbf{g}$ is specified analytically, we recreate the example posed in  \cite{weisman2016SolutionLiouvilleEquation}, which considers the evolution of the semi-major axis $a$ and mean anomaly $M$ of a satellite under two-body dynamics.
Because all of the other orbital elements are constants of motion and the mean anomaly only depends on the semi-major axis, the joint distribution of semi-major axis and mean anomaly are of primary interest.
These two variables evolve according to
\begin{align}
    [a(t)\quad M(t)]^T&=\mathbf{g}([a(t_0)\quad M(t_0)]^T)\\
    &=\left[a(t_0)\quad M(t_0)+\sqrt{\frac{\mu}{a(t_0)^3}}(t-t_0)\right]^T
\end{align}
The joint distribution of the semi-major axis and mean anomaly at the initial epoch is taken as the Gaussian
\begin{align}
    p(\mathbf{x}(t_0)) &= \mathcal{N}(\mathbf{x}(t_{0}); \,\boldsymbol{\mu}_{0}, \, \mathbf{P}_{0})\\
\boldsymbol{\mu}_{0} &= \begin{bmatrix}
\mu_{a} & \mu_{M_{0}}
\end{bmatrix}^{T}\\
    \mathbf{P}_{0} &= \operatorname{diag}([\sigma_{a_{0}}^2 \quad \sigma_{M_{0}}^2])
\end{align}
where $\mu_{a_{0}}=1.4322 \,[\textrm{Earth Radii (ER)}]$, $\mu_{M_{0}}= 0 \, [\textrm{rad}]$, $\sigma_{a}=0.25 \, [\textrm{ER}]$, and $\sigma_{M_{0}}=0.02 \, [\textrm{rad}]$.
Each splitting method is limited to maximum splitting recursion depth of four, such that the final mixtures are composed of 81 mixands.
Because in this example $\mathbf{g}$ is bijective with determinant of the Jacobian equal to one, the true \ac{pdf} at an arbitrary later time $t$ is obtained by the change of variables formula and given by
\begin{equation}
p(\mathbf{x}(t))\!=\!\frac{1}{K} \exp\left(\!{-\frac{\left(a-\mu_{a_{0}}\right)^2}{2\sigma_{a_{0}}^2}-\frac{\left(M\!-\!\sqrt{\frac{\mu}{a^3}} t\!-\!\mu_{M_{0}}\right)^2}{2\sigma_{M_{0}}^2}}\right)
\end{equation}
where the normalization constant $K=2\pi \sigma_{a_{0}} \sigma_{M_{0}}$ \cite{weisman2016SolutionLiouvilleEquation}.
The true \ac{pdf} after two orbital periods of the mean orbit (around 2.4 hours) is shown in Figure~\ref{fig:twobody:truth}.
The approximation accuracy of each method, as measured by the \ac{nise}, is tabulated in Table~\ref{tab:twobody_results}.
The resulting \ac{pdf} are shown in Fig.~\ref{fig:two_body}.
\begin{table}[htbp]
  \caption{Two-body motion example.}
  \label{tab:twobody_results}
  \centering
    \begin{tabular}{lr}
      \toprule[2pt]%
    Method & NISE$\downarrow$\\
      \midrule
      \begin{filecontents*}{twobody_results.csv}
method,NISE
maxvar   , 0.008054 
ALoDT    , 0.008054 
FOS      , 0.581873 
SOS      , 0.008054 
SOLC     , 0.008054 
SADL     , 0.008054 
USFOS    , 0.008931 
USSOLC   , 0.008054 
SAFOS    , 0.008569 
SASOS    , 0.008054 
WUSSOS   , 0.008054 
WUSSOLC  , 0.008054 
WUSSADL  , 0.008054 
WSASOS   , 0.008054 
\end{filecontents*}
\csvreader[head to column names]{twobody_results.csv}{}%
      {\method & \NISE\\}\\[-1em]
      \bottomrule
    \end{tabular}
\end{table}

In this example, all of the methods produce nearly identical results with the exception of the \ac{fos} method.
Whereas the second-order stretching direction method picks exactly the semi-major axis direction for splitting, first-order splitting chooses the direction along the unit vector $[0.997158 \quad 0.0753323]$.
While this direction is very nearly aligned with the semi-major axis direction, a small difference in direction makes a very large difference in the resulting placement of the mixands given the much higher level of uncertainty in semi-major axis than in mean anomaly.
The square root of the inverse precision along the semi-major axis direction for the initial distribution is $0.25\,[\textrm{ER}]$, but the same quantity associated with the \ac{fos} splitting direction is only $0.182\,[\textrm{ER}]$.
As such, the mixands at each splitting step will be placed around two thirds of the distance from the original mean as they would be if split exactly along the semi-major axis direction.
With the mixand spread less far away from the original mean, the resulting distribution will be more Gaussian than a distribution that results from mixands that are closer to the mean but in approximately the same direction.

The uncertainty scaled first-order stretching method also does not align exactly with the semi-major axis direction, but is even closer to the semi-major axis direction, making the splitting nearly indistinguishable from the second-order splitting method.
Notably, all other methods besides first-order stretching perform nearly identically.
This example highlights the need to account for the initial uncertainty distribution in choosing the splitting directions, though it so happens that the nonlinear splitting methods happen to work in this case even when failing to consider the initial uncertainty distribution.

\subsection{THREE-BODY MOTION}
\begin{figure}[htpb]
  \centering
  \includegraphics[width=0.99\linewidth]{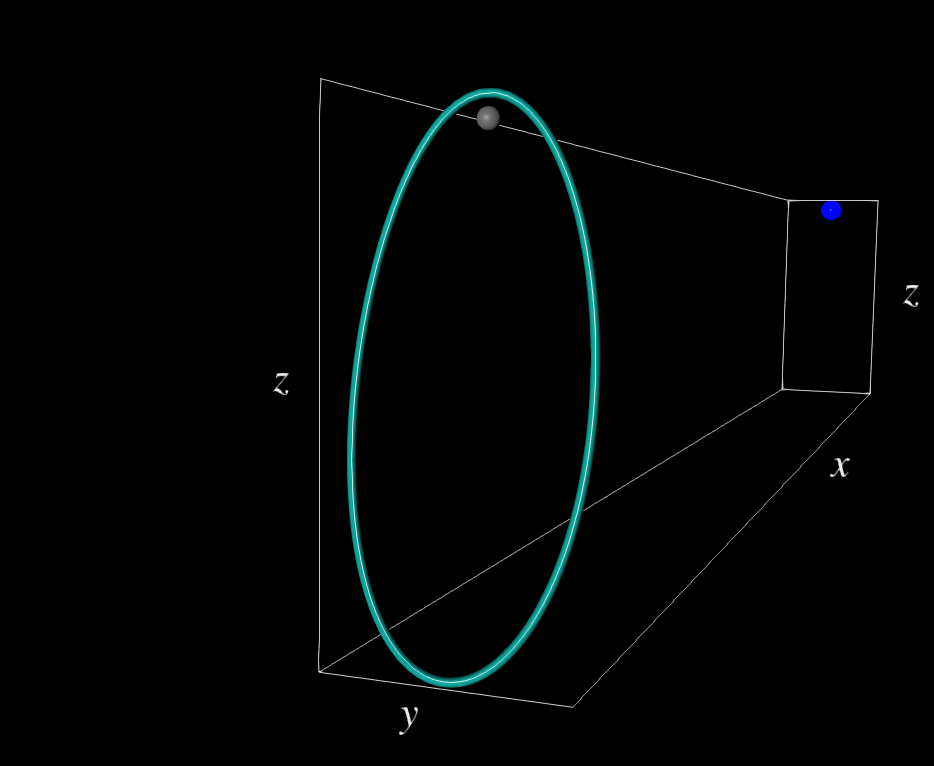}
  \caption{Full NRHO considered in three-body uncertainty propagation application.}%
  \label{fig:nrho}
\end{figure}

The equations of motion for the circular restricted three-body problem are given in the synodic frame as
\begin{align}
    \frac{\mathrm{d}}{\mathrm{d}t}\mathbf{x}&=\mathbf{F(x)}\\
    \mathbf{F(x)}&=\begin{bmatrix} \dot{x}& \dot{y}& \dot{z}& 2\dot{y}+\frac{\partial \overline{U}}{\partial x}& -2\dot{x}+\frac{\partial \overline{U}}{\partial y}& \frac{\partial \overline{U}}{\partial z}\end{bmatrix}^T
    \label{eqn:cr3bp}
\end{align}
where $\overline{U}(x,y,z)=\dfrac{1-\mu^*}{||\mathbf{r}_1||}+\dfrac{\mu^*}{||\mathbf{r}_2||}+\dfrac{x^2+y^2}{2}$ is the effective potential, the reduced mass is defined as $\mu^*=\dfrac{m_2}{m_1+m_2}$ for the two primary bodies with masses $m_1,m_2$ respectively.
The primary body with greater mass is given index $1$ so that $m_1\geq m_2$.
Both masses are located along the $x$-axis at $[-\mu^*,0,0]$ and $[1-\mu^*, 0, 0]$ with respect to their common barycenter at the origin.
The position of the satellite of interest with respect to the primary and secondary bodies is given by $\mathbf{r}_1$ and $\mathbf{r}_2$ respectively  \cite{koon2000dynamical}.
The reference orbit used in the following sections comes from the proposed NASA Gateway orbit \cite{NationalAA2019} and is shown in Figure~\ref{fig:nrho}.
The initial conditions and mass parameter used for the orbit are
\begin{gather*}
    \mu=1.0/(81.30059 + 1.0), \quad x_0=1.022022,\\z_0 = -0.182097, \quad \dot{y}_0 = -0.103256
\end{gather*}
in nondimensional units with other initial coordinates equal to zero.
This initial condition coincides with the apolune of the orbit.
The period of the orbit is approximately $1.511111$ nondimensional time units where $2\pi$ time units correspond to the period of revolution for the Earth-Moon system.
In this example, the nonlinear function $\mathbf{g}$ represents the flow of the circular restricted three-body dynamics $\varphi_t$ for some fixed value of the time-of-flight $t$.
The flow map associated with the dynamical system in~\eqref{eqn:cr3bp} is defined such that
\begin{equation}
\frac{\mathrm{d}}{\mathrm{d}t}\varphi_t(\mathbf{x}_0)=\mathbf{F}(\varphi_t(\mathbf{x}_0)), \quad \varphi_0(\mathbf{x}_0)=\mathbf{x}_0
\end{equation}

The Jacobian and the second-order partial derivative tensor of the flow map around some reference trajectory are known in the astrodynamics literature as the state transition matrix and the second-order state transition tensor respectively \cite{park2006nonlinear}.
While it can be costly to compute the state transition tensors by integrating the variational equations associated with them, there exist efficient methods for precomputing and then interpolating the state transition tensors along an a priori known reference trajectory \cite{cunningham2023interpolated,kulik2023state,cunningham2024spice}, as well as approximate computation methods that that rely on the potential low-rank qualities of the state transition tensors \cite{boone2023directional,boone2024efficient}.

In this example, we employ an initial Gaussian distribution with mean equal to the Gateway initial conditions and covariance given by
\begin{equation}
    \mathbf{P}_x=10^{-8}\mathrm{diag}([1,0,1,0,0,0])+10^{-10}\mathbf{I}_6
\end{equation}
in nondimensional canonical units.
The 1-sigma distances are on the order of $40 \, [\textrm{km}]$ along the $x$ and $z$ directions, $4\, [\textrm{km}]$ along the $y$ direction, and $0.01 \, [\textrm{m/s}]$ in each velocity direction.
We propagate for half of an orbital period which amounts to just under one quarter of a month, and leaves the mean trajectory at its perilune.
The flow of the dynamics between apolune and perilune is highly nonlinear \cite{kulik2024applications} which makes it a stressing case for uncertainty propagation.
The recursive split operation is limited to a recursion depth of three, resulting in a final mixture with $27$ mixands.

\begin{table*}[htbp]
  \caption{Three-body motion example.}
  \label{tab:cislunar_results}
  \centering
    \begin{tabular}{lrrrr}
      \toprule[2pt]%
      Method & ELK$\uparrow$ & MaDEM$\downarrow$ & MCR$\downarrow$ & CVM norm$\downarrow$\\
      \midrule
      \begin{filecontents*}{cislunar_results.csv}
method  , ELK        , MaDEM  , CvMnorm , MCR
maxvar  , 6.3941e+22 , 0.4127 , 314.40  , 2.081
ALoDT   , 8.8427e+22 , 0.2822 , 94.28   , 1.309
FOS     , 4.5372e+22 , 2.3745 , 1156    , 14.30
SOS     , 4.5369e+22 , 2.3798 , 1157    , 14.34
SOLC    , 4.5227e+22 , 2.3869 , 1158    , 14.36
SADL    , 8.8241e+22 , 0.2224 , 75.25   , 1.287
USFOS   , 8.7855e+22 , 0.0891 , 43.09   , 1.267
USSOLC  , 8.7824e+22 , 0.0888 , 42.97   , 1.267
SAFOS   , 6.7044e+22 , 0.1862 , 136.5   , 1.450
SASOS   , 8.7824e+22 , 0.0889 , 42.96   , 1.267
WUSSOS  , 8.7824e+22 , 0.0888 , 42.97   , 1.267
WUSSOLC , 8.7824e+22 , 0.0888 , 42.97   , 1.267
WUSSADL , 8.8401e+22 , 0.2270 , 76.84   , 1.287
WSASOS  , 8.7831e+22 , 0.0889 , 42.99   , 1.267
\end{filecontents*}
\csvreader[head to column names]{cislunar_results.csv}{}%
      {\method & \ELK & \MaDEM & \MCR & \CvMnorm\\}\\[-1em]%
      \bottomrule
    \end{tabular}
\end{table*}

\begin{figure*}
\newcommand{\figwidthfactor}{0.3}
\newcommand{\example}{cislunar}
\newcommand{\plotsuffix}{_0_1}
\centering
\begin{subcaptiongroup}
  \subcaptionlistentry{throwaway}
  \label{fig:\example:truth}
  \begin{tikzpicture}
    \node[anchor=north west] (img) at (0,0){\includesvg[width=\figwidthfactor\linewidth]{graphics/\example truth_hist\plotsuffix.svg}};
    \node at (0.4,-0.6) {\captiontext*{}};
  \end{tikzpicture}
  \subcaptionlistentry{throwaway}
  \label{fig:\example:truth_scatter}
  \begin{tikzpicture}
    \node[anchor=north west] (img) at (0,0){\includesvg[width=\figwidthfactor\linewidth]{graphics/\example truth_scatter\plotsuffix.svg}};
    \node at (0.4,-0.2) {\captiontext*{}};
  \end{tikzpicture}
  \subcaptionlistentry{throwaway}
  \label{fig:\example:variance}
  \begin{tikzpicture}
    \node[anchor=north west] (img) at (0,0){\includesvg[width=\figwidthfactor\linewidth]{graphics/\example variance\plotsuffix.svg}};
    \node at (0.4,-0.2) {\captiontext*{}};
  \end{tikzpicture}
  \subcaptionlistentry{throwaway}
  \label{fig:\example:fos}
  \begin{tikzpicture}
    \node[anchor=north west] (img) at (0,0){\includesvg[width=\figwidthfactor\linewidth]{graphics/\example FOS\plotsuffix.svg}};
    \node at (0.4,-0.6) {\captiontext*{}};
  \end{tikzpicture}
  \subcaptionlistentry{throwaway}
  \label{fig:\example:safos}
  \begin{tikzpicture}
    \node[anchor=north west] (img) at (0,0){\includesvg[width=\figwidthfactor\linewidth]{graphics/\example SAFOS\plotsuffix.svg}};
    \node at (0.4,-0.6) {\captiontext*{}};
  \end{tikzpicture}
  \subcaptionlistentry{throwaway}
  \label{fig:\example:usfos}
  \begin{tikzpicture}
    \node[anchor=north west] (img) at (0,0)
{\includesvg[width=\figwidthfactor\linewidth]{graphics/\example USFOS\plotsuffix.svg}};
    \node at (0.4,-0.6) {\captiontext*{}};
  \end{tikzpicture}
  \subcaptionlistentry{throwaway}
  \label{fig:\example:wussadl}
  \begin{tikzpicture}
    \node[anchor=north west] (img) at (0,0)
{\includesvg[width=\figwidthfactor\linewidth]{graphics/\example WUSSADL\plotsuffix.svg}};
    \node at (0.4,-0.6) {\captiontext*{}};
  \end{tikzpicture}
  \subcaptionlistentry{throwaway}
  \label{fig:\example:alodt}
  \begin{tikzpicture}
    \node[anchor=north west] (img) at (0,0)
{\includesvg[width=\figwidthfactor\linewidth]{graphics/\example ALoDT\plotsuffix.svg}};
    \node at (0.4,-0.6) {\captiontext*{}};
  \end{tikzpicture}
\end{subcaptiongroup}
\captionsetup{subrefformat=parens}
\caption{Three-body nonlinear-mapped \subref{fig:cislunar:truth} and \subref{fig:cislunar:truth_scatter} truth and nonlinear-mapped split densities using \subref{fig:cislunar:variance} max-variance, \subref{fig:cislunar:fos} \ac{fos} (identical to \ac{sos} and \ac{solc}), \subref{fig:cislunar:safos} \ac{safos}, \subref{fig:cislunar:usfos} \ac{usfos} (identical to \ac{ussolc}, \ac{sasos}, \ac{wussos}, \ac{wussolc}, and \ac{wsasos}),  \subref{fig:cislunar:wussadl} \ac{wussadl} (identical to \ac{sadl}), and \subref{fig:cislunar:alodt} \ac{alodt}.}

\label{fig:cislunar_split_transformed}
\end{figure*}
The approximation accuracy of each method, as measured by the \ac{nise}, is tabulated in Table~\ref{tab:cislunar_results}.
Fig.~\ref{fig:cislunar_split_transformed} shows the $x$-$y$ marginals of the final distribution propagated using a Monte Carlo approach as well as the \ac{gm} splitting methods described in this paper.
The distribution is highly non-Gaussian.
We see that all of the methods that do not incorporate initial uncertainty (\ac{fos}, \ac{sos}, \ac{solc}) perform poorly according to all four methods of assessing the quality of the uncertainty propagation (\ac{elk}, \ac{MaDEM}, \ac{mcr}, and \ac{cvm} norm).
\Ac{sadl} and \ac{alodt} implicitly involve information about the initial uncertainty since the statistical linearization relies on sigma points that are chosen according to the initial distribution, so it works better than the other three methods that do not explicitly account for the initial uncertainty in choosing the splitting direction.
\Ac{wussadl} performs roughly the same as \ac{sadl} under each metric since the problem employs nondimensional coordinates, and finally the six methods \ac{usfos}, \ac{wussos}, \ac{ussolc}, \ac{wussolc}, \ac{sasos}, and \ac{wsasos} all perform roughly the same in this instance under each metric.
Interestingly, \ac{wussadl} and \ac{alodt} perform best in terms of \ac{elk} as compared with the other six front-runner methods, but worse under comparison of the moments with the Monte Carlo and the \ac{cvm} norm.
By visual inspection, the contours of equal probability density of the distribution propagated with \ac{wussadl} are less smooth than the contours associated with the other four front-runners.
Variance only (maxvar) splitting is even less performant than the \ac{sadl} based methods under all metrics and visual inspection, but is far better than the heuristics informed only by the dynamics without the uncertainty, such as \ac{fos}, \ac{sos}, and \ac{solc}.
An important takeaway is that the performance of the uncertainty weighted \ac{fos} and \ac{sos} splitting methods are nearly identical because the maximal linear and nonlinear stretching directions align very closely under these dynamics and this propagation time \cite{boone2023directional}.
As it is the simplest method and performs among the best in terms of the metrics of comparison with Monte Carlo, \ac{usfos} is likely the method of choice for three-body dynamics, at least for propagation over relatively long timescales where the stretching directions are expected to align.

\section{CHOOSING A SPLITTING CRITERION}
\label{sec:discussion}
We have presented a potentially overwhelming number of methods for choosing the splitting direction.
None of these methods performs uniformly best in terms of approximation error for all cases studied, making it potentially difficult to choose a splitting method.
In order to help practitioners select a method that works well for their application, Table~\ref{table:summary} qualitatively summarizes the performance of each method according to a number of factors: interpretability of the objective function as a meaningful splitting criterion, performance in the measurement scenario, performance for 2- and 3-body dynamics propagation, and finally, the anticipated ease of implementation of the method.
Computational performance is omitted from this list as it is highly application dependent.
For instance, access to the first- and second-order partial derivatives of the function may or may not have already been computed for another purpose, and otherwise may or may not be a dominating cost for the computation depending on the nature of the nonlinear function $\mathbf{g}$.

We note that \ac{wsasos}, \ac{wussadl}, and \ac{wussolc} are all good general purpose methods, albeit with a more technically-difficult implementation.
For astrodynamics problems, \ac{usfos} is an easy to implement and highly performant method even if the splitting criterion associated with it is less interpretable than others and lacks scale independence under coordinate changes.
In these cases, \ac{usfos} should be employed in nondimensional coordinates to perform best.
\Ac{safos} is a good choice for general purpose use when ease of implementation is an important consideration.
\ac{wussos} performs on the same level (and sometimes slightly better) as \ac{wussolc}, which can be viewed as a relaxation of \ac{wussos}.
\ac{wussolc} is simpler to implement successfully and can be computed deterministically without random initial guesses, so \ac{wussolc} or \ac{wsasos} are generally to be preferred.
\begin{table*}[htbp]
  \caption{Splitting direction selection summary.}
  \centering
    \begin{tabular}{lrrrr}
      \toprule[2pt]%
      Method & Interpretable Criterion & Measurement Performance	& Dynamics Performance	& Implementation Ease\\
      \midrule
      \begin{filecontents*}{method_comparison.csv}
method,Interpretable,Measurement,Dynamics,Implementation
maxvar,Yes,High,Low,High
USFOS,No,Low,High,High
SAFOS,No,High,Mid,High
SASOS,No,High,High,Mid
WUSSOS,Yes,High,High,Low
WUSSOLC,Yes,High,High,Mid
WUSSADL,Yes,High,Mid,High
WSASOS,Yes,High,High,Mid
\end{filecontents*}
\csvreader[head to column names]{method_comparison.csv}{}%
      {\method & \Interpretable & \Measurement & \Dynamics & \Implementation\\}\\[-1em]%
      \bottomrule
      \label{table:summary}
    \end{tabular}
\end{table*}

\section{CONCLUSION}
\label{sec:conclusion}
The contributions of this work are two-fold.
First, a moment-matching Gaussian mixture splitting method is presented that allows for splitting along arbitrary directions with potentially nonuniform mixand variances along the splitting direction.
Second, nine novel methods are presented for selecting the splitting direction of a Gaussian mixture for uncertainty propagation through a nonlinear function.
Splitting direction selection is informed by the degree of uncertainty in each direction as well as the properties of the function through which the distribution is propagated.
A comparison of these methods was performed in the context of three examples that include nonlinear dynamics as well as nonlinear measurement functions.
This analysis yielded three general findings.
First, accounting for the degree of initial uncertainty in each direction tends to yield Gaussian mixtures that better approximate the true distribution.
Second, metrics that characterize an uncertainty weighted measure of the level of nonlinearity work best among all of the methods considered.
Third, methods that rely on a function's linear stretching properties can work well in the case of the flow of two- and three-body dynamics, but do not work in the case of measurement functions and some other classes of dynamical systems.

\section*{APPENDIX}
\subsection{Positive Definiteness of Mixand Covariance}
\label{sec:posdef_app}
In general, a downdate of a symmetric positive definite matrix $\mathbf{A}$ by the rank-1 matrix $-\alpha \hat{\mathbf{v}}\hat{\mathbf{v}}^T$ (where $\alpha$ is some positive scalar, and $\hat{\mathbf{v}}$ is a unit vector with the same dimensions as $\mathbf{A}$) gives a positive semi-definite result if and only if the change is sufficiently small as measured by the parameter $\alpha$  \cite{bernstein2009matrix}.
In particular,
\begin{align}
    &0\leq\mathbf{A}-\alpha\hat{\mathbf{v}}\hat{\mathbf{v}}^T\\
    \iff&\alpha\leq\frac{1}{\hat{\mathbf{v}}^T\mathbf{A}^{-1}\hat{\mathbf{v}}}=\alpha^*
\end{align}
This can be derived by examining the values of $\alpha$ such that $\mathbf{A}-\alpha\hat{\mathbf{v}}\hat{\mathbf{v}}^T$ is singular and there exists $\mathbf{x}\neq 0$ such that
\begin{equation}
    (\mathbf{A}-\alpha\hat{\mathbf{v}}\hat{\mathbf{v}}^T)\mathbf{x}=0
\end{equation}
These correspond to the reciprocal of eigenvalues $\lambda$ of the generalized eigenvalue problem associated with the matrix pencil $(\hat{\mathbf{v}}\hat{\mathbf{v}}^T, \mathbf{A})$:
\begin{equation}
    \hat{\mathbf{v}}\hat{\mathbf{v}}^T\mathbf{x}=\lambda\mathbf{A}\mathbf{x}
\end{equation}
where $\lambda=1/\alpha$.
Because both matrices in the pencil are symmetric positive semi-definite, the eigenvectors form a $\hat{\mathbf{v}}\hat{\mathbf{v}}^T$-orthogonal set, meaning that for two eigenvectors $\mathbf{x}_i,\mathbf{x}_j$
\begin{equation}
    \mathbf{x}_i^T\hat{\mathbf{v}}\hat{\mathbf{v}}^T\mathbf{x}_j = \delta_{i,j}
\end{equation}
where $\delta_{i,j}$ is the Kronecker delta.
One such eigenvector is $\mathbf{A}^{-1}\hat{\mathbf{v}}$ with corresponding eigenvalue $\lambda=1/\alpha^*$.
As a result of the $\hat{\mathbf{v}}\hat{\mathbf{v}}^T$-orthogonality of the eigenvectors, any other eigenvectors must be orthogonal to $\hat{\mathbf{v}}$ and thus correspond to zero eigenvalues.
This means that the only value of $\alpha$ that renders the matrix resulting from the rank-1 downdate singular is $\alpha^*$.
For $\alpha$ smaller than $\alpha^*$, the rank-1 downdate is positive definite, and above this threshold, the downdate is no-longer positive definite.
The eigenvector $\mathbf{A}^{-1}\hat{\mathbf{v}}$ gives the direction in which the downdated matrix becomes singular.
In the case of the downdated original covariance, this is the direction in which the resulting mixand covariance becomes zero when the value of $\alpha$ is equal to $\alpha^*$.
In the context of determining whether $\bar{\mathbf{P}}$ is positive semi-definite
\begin{equation}
    \alpha = \sum_{i=1}^{L} w_i\Vert\boldsymbol{\mu}_i-\boldsymbol{\mu}\Vert^2
\end{equation}
the matrix $\mathbf{A}=\mathbf{P}$, and $\hat{\mathbf{v}}=\hat{\mathbf{x}}^*$.
Thus, valid mixand covariances are only possible if the univariate splitting and original covariance are compatible:
\begin{equation}
    \label{eq:spd-cond}
    \sum_{i=1}^{L} w_i\Vert\boldsymbol{\mu}_i-\boldsymbol{\mu}\Vert^2 \leq \frac{1}{(\hat{\mathbf{x}}^*)^T\mathbf{P}^{-1}\hat{\mathbf{x}}^*}
\end{equation}
Equation~\eqref{eq:spd-cond} is a stronger condition than the expression
\begin{equation}
    \label{eq:univariate-cond}
    (\hat{\mathbf{x}}^*)^T\mathbf{P}\hat{\mathbf{x}}^*\geq\alpha
\end{equation}
which states that the variance of the weighted means of the mixands must be less than the directional variance $(\hat{\mathbf{x}}^*)^T\mathbf{P}\hat{\mathbf{x}}^*$ of the original distribution.
The stronger nature of~\eqref{eq:spd-cond} relative to~\eqref{eq:univariate-cond} comes from the fact that, for all vectors $\mathbf{v}$ and symmetric positive definite~$\mathbf{A}$,
\begin{align}
    \mathbf{v}^T\mathbf{A}\mathbf{v}\geq\frac{1}{\mathbf{v}^T\mathbf{A}^{-1}\mathbf{v}}
\end{align}
which is evident from the eigendecomposition of $\mathbf{A}$ and the inequality of similarly weighted harmonic and arithmetic means.

\bibliographystyle{IEEEtran}
\bibliography{mybib}

\begin{thebibliography}{10}
\providecommand{\url}[1]{#1}
\csname url@samestyle\endcsname
\providecommand{\newblock}{\relax}
\providecommand{\bibinfo}[2]{#2}
\providecommand{\BIBentrySTDinterwordspacing}{\spaceskip=0pt\relax}
\providecommand{\BIBentryALTinterwordstretchfactor}{4}
\providecommand{\BIBentryALTinterwordspacing}{\spaceskip=\fontdimen2\font plus
\BIBentryALTinterwordstretchfactor\fontdimen3\font minus \fontdimen4\font\relax}
\providecommand{\BIBforeignlanguage}[2]{{%
\expandafter\ifx\csname l@#1\endcsname\relax
\typeout{** WARNING: IEEEtran.bst: No hyphenation pattern has been}%
\typeout{** loaded for the language `#1'. Using the pattern for}%
\typeout{** the default language instead.}%
\else
\language=\csname l@#1\endcsname
\fi
#2}}
\providecommand{\BIBdecl}{\relax}
\BIBdecl

\bibitem{julier2004unscented}
S.~J. Julier and J.~K. Uhlmann, ``Unscented filtering and nonlinear estimation,'' \emph{Proceedings of the IEEE}, vol.~92, no.~3, pp. 401--422, 2004.

\bibitem{alspach1972nonlinear}
D.~Alspach and H.~Sorenson, ``Nonlinear bayesian estimation using {G}aussian sum approximations,'' \emph{IEEE transactions on automatic control}, vol.~17, no.~4, pp. 439--448, 1972.

\bibitem{anderson1979optimal}
B.~Anderson and J.~B. Moore, \emph{Optimal Filtering}.\hskip 1em plus 0.5em minus 0.4em\relax Prentice-Hall Information and System Sciences Series, Englewood Cliffs: Prentice-Hall, 1979.

\bibitem{demars2013entropy}
K.~J. DeMars, R.~H. Bishop, and M.~K. Jah, ``Entropy-based approach for uncertainty propagation of nonlinear dynamical systems,'' \emph{Journal of Guidance, Control, and Dynamics}, vol.~36, no.~4, pp. 1047--1057, 2013.

\bibitem{vittaldev2016spacecraft}
V.~Vittaldev, R.~P. Russell, and R.~Linares, ``Spacecraft uncertainty propagation using {G}aussian mixture models and polynomial chaos expansions,'' \emph{Journal of Guidance, Control, and Dynamics}, vol.~39, no.~12, pp. 2615--2626, 2016.

\bibitem{faubel2009split}
F.~Faubel, J.~McDonough, and D.~Klakow, ``The split and merge unscented {G}aussian mixture filter,'' \emph{IEEE Signal Processing Letters}, vol.~16, no.~9, pp. 786--789, 2009.

\bibitem{faubel2010further}
F.~Faubel and D.~Klakow, ``Further improvement of the adaptive level of detail transform: Splitting in direction of the nonlinearity,'' in \emph{2010 18th European Signal Processing Conference}.\hskip 1em plus 0.5em minus 0.4em\relax IEEE, 2010, pp. 850--854.

\bibitem{leutnant2011versatile}
V.~Leutnant, A.~Krueger, and R.~Haeb-Umbach, ``A versatile {G}aussian splitting approach to non-linear state estimation and its application to noise-robust asr.'' in \emph{Interspeech}, 2011, pp. 1641--1644.

\bibitem{huber2011AdaptiveGaussianMixture}
M.~F. Huber, ``Adaptive {{Gaussian}} mixture filter based on statistical linearization,'' in \emph{14th {{International Conference}} on {{Information Fusion}}}, 2011, pp. 1--8.

\bibitem{jones2024physics}
B.~A. Jones, ``Physics-informed domain splitting for orbit uncertainty propagation,'' in \emph{AIAA SCITECH 2024 Forum}, 2024, p. 0203.

\bibitem{tuggle2018automated}
K.~Tuggle and R.~Zanetti, ``Automated splitting {G}aussian mixture nonlinear measurement update,'' \emph{Journal of Guidance, Control, and Dynamics}, vol.~41, no.~3, pp. 725--734, 2018.

\bibitem{tuggle2020model}
K.~E. Tuggle, ``Model selection for {G}aussian mixture model filtering and sensor scheduling,'' Ph.D. dissertation, 2020.

\bibitem{legrand2022SplitHappensImprecise}
K.~A. LeGrand and S.~Ferrari, ``Split {{Happens}}! {{Imprecise}} and {{Negative Information}} in {{Gaussian Mixture Random Finite Set Filtering}},'' \emph{Journal of Advances in Information Fusion}, vol.~17, no.~2, pp. 78--96, Dec. 2022.

\bibitem{dunik2018directional}
J.~Dun{\'\i}k, O.~Straka, B.~Noack, J.~Steinbring, and U.~D. Hanebeck, ``Directional splitting of {G}aussian density in non-linear random variable transformation,'' \emph{IET Signal Processing}, vol.~12, no.~9, pp. 1073--1081, 2018.

\bibitem{gutierrez2024classifying}
J.~Gutierrez, K.~Hill, E.~L. Jenson, D.~J. Scheeres, J.~C. Bruer, and R.~D. Coder, ``Classifying state uncertainty for {E}arth-{M}oon trajectories,'' \emph{The Journal of the Astronautical Sciences}, vol.~71, no.~3, p.~29, 2024.

\bibitem{kulik2024applications}
J.~Kulik, C.~Orton-Urbina, M.~Ruth, and D.~Savransky, ``Applications of induced tensor norms to guidance navigation and control,'' \emph{arXiv preprint arXiv:2408.15362}, 2024.

\bibitem{jenson2023semianalytical}
E.~L. Jenson and D.~J. Scheeres, ``Semianalytical measures of nonlinearity based on tensor eigenpairs,'' \emph{Journal of Guidance, Control, and Dynamics}, vol.~46, no.~4, pp. 638--653, 2023.

\bibitem{sarkka2023bayesian}
S.~S{\"a}rkk{\"a} and L.~Svensson, \emph{Bayesian filtering and smoothing}.\hskip 1em plus 0.5em minus 0.4em\relax Cambridge university press, 2023, vol.~17.

\bibitem{NationalAA2019}
``National aeronautics and space administration (nasa) white paper: Gateway destination orbit model: A continuous 15 year nrho reference trajectory,'' 2019.

\bibitem{golub2013matrix}
G.~H. Golub and C.~F. Van~Loan, \emph{Matrix computations}.\hskip 1em plus 0.5em minus 0.4em\relax JHU press, 2013.

\bibitem{hanebeck2003ProgressiveBayesNew}
U.~D. Hanebeck, K.~Briechle, and A.~Rauh, ``Progressive {{Bayes}}: {{A New Framework}} for {{Nonlinear State Estimation}},'' in \emph{Proceedings of {{SPIE AeroSense Symposium}}}, vol. 5099, 2003, pp. 256--267.

\bibitem{bunch1978rank}
J.~R. Bunch, C.~P. Nielsen, and D.~C. Sorensen, ``Rank-one modification of the symmetric eigenproblem,'' \emph{Numerische Mathematik}, vol.~31, no.~1, pp. 31--48, 1978.

\bibitem{bojanczyk1987note}
A.~W. Bojanczyk, R.~Brent, P.~Van~Dooren, and F.~De~Hoog, ``A note on downdating the {C}holesky factorization,'' \emph{SIAM journal on scientific and statistical computing}, vol.~8, no.~3, pp. 210--221, 1987.

\bibitem{cmes}
R.~P.~R. V.~Vittaldev, ``Multidirectional {G}aussian mixture models for nonlinear uncertainty propagation,'' \emph{Computer Modeling in Engineering \& Sciences}, vol. 111, no.~1, pp. 83--117, 2016.

\bibitem{calkins2024}
G.~E. Calkins, J.~W. McMahon, and D.~C. Woffinden, ``Efficient higher-order analytical covariance analysis for aerocapture,'' in \emph{Astrodynamics Specialist Conference, 2024}, 2024, pp. AAS 24--241.

\bibitem{losacco2024low}
M.~Losacco, A.~Foss{\`a}, and R.~Armellin, ``Low-order automatic domain splitting approach for nonlinear uncertainty mapping,'' \emph{Journal of Guidance, Control, and Dynamics}, pp. 1--20, 2024.

\bibitem{boone2023directional}
S.~Boone and J.~McMahon, ``Directional state transition tensors for capturing dominant nonlinear effects in orbital dynamics,'' \emph{Journal of Guidance, Control, and Dynamics}, vol.~46, no.~3, pp. 431--442, 2023.

\bibitem{jenson2024bounding}
E.~L. Jenson and D.~J. Scheeres, ``Bounding nonlinear stretching about spacecraft trajectories using tensor eigenpairs,'' \emph{Acta Astronautica}, vol. 214, pp. 159--166, 2024.

\bibitem{park2006nonlinear}
R.~S. Park and D.~J. Scheeres, ``Nonlinear mapping of {G}aussian statistics: theory and applications to spacecraft trajectory design,'' \emph{Journal of guidance, Control, and Dynamics}, vol.~29, no.~6, pp. 1367--1375, 2006.

\bibitem{rasotto2016differential}
M.~Rasotto, A.~Morselli, A.~Wittig, M.~Massari, P.~Di~Lizia, R.~Armellin, C.~Valles, and G.~Ortega, ``Differential algebra space toolbox for nonlinear uncertainty propagation in space dynamics,'' 2016.

\bibitem{kolda2011shifted}
T.~G. Kolda and J.~R. Mayo, ``Shifted power method for computing tensor eigenpairs,'' \emph{SIAM Journal on Matrix Analysis and Applications}, vol.~32, no.~4, pp. 1095--1124, 2011.

\bibitem{qi2005eigenvalues}
L.~Qi, ``Eigenvalues of a real supersymmetric tensor,'' \emph{Journal of symbolic computation}, vol.~40, no.~6, pp. 1302--1324, 2005.

\bibitem{kumar2022splitting}
A.~Kumar, \emph{Splitting {G}aussian Densities to Minimize Variance Along a Direction of Nonlinearity}.\hskip 1em plus 0.5em minus 0.4em\relax Portland State University, 2022.

\bibitem{iannamorelli2024AdaptiveGaussianMixture}
J.~Iannamorelli and K.~LeGrand, ``Adaptive {{Gaussian Mixture Filtering}} for {{Multi-Sensor Maneuvering Cislunar Space Object Tracking}},'' \emph{(in press) Journal of the Astronautical Sciences}, 2024.

\bibitem{williams2003GaussianMixtureReduction}
J.~L. Williams, ``Gaussian {{Mixture Reduction}} of {{Tracking Multiple Maneuvering Targets}} in {{Clutter}},'' M.{{S}}. {{Thesis}}, Air Force Institute of Technology, 2003.

\bibitem{darling1957KolmogorovSmirnovCramervonMises}
D.~A. Darling, ``The {{Kolmogorov-Smirnov}}, {{Cramer-von Mises Tests}},'' \emph{The Annals of Mathematical Statistics}, vol.~28, no.~4, pp. 823--838, Dec. 1957.

\bibitem{jebara2004ProbabilityProductKernels}
T.~Jebara, R.~Kondor, and A.~Howard, ``Probability product kernels,'' \emph{Journal of Machine Learning Research}, vol.~5, pp. 819--844, Dec. 2004.

\bibitem{faubel2010FurtherImprovementAdaptive}
F.~Faubel and D.~Klakow, ``Further improvement of the adaptive level of detail transform: {{Splitting}} in direction of the nonlinearity,'' in \emph{2010 18th {{European Signal Processing Conference}}}, Aug. 2010, pp. 850--854.

\bibitem{julier2002ScaledUnscentedTransformation}
S.~Julier, ``The scaled unscented transformation,'' in \emph{Proceedings of the 2002 {{American Control Conference}} ({{IEEE Cat}}. {{No}}.{{CH37301}})}.\hskip 1em plus 0.5em minus 0.4em\relax Anchorage, AK, USA: IEEE, 2002, pp. 4555--4559 vol.6.

\bibitem{weisman2016SolutionLiouvilleEquation}
R.~Weisman, M.~Majji, and K.~T. Alfriend, ``Solution of {L}iouville's equation for uncertainty characterization of the main problem in satellite,'' \emph{Tech Science Press CMES}, vol. 111, no.~3, pp. 269--304, 2016.

\bibitem{koon2000dynamical}
W.~S. Koon, M.~W. Lo, J.~E. Marsden, and S.~D. Ross, ``Dynamical systems, the three-body problem and space mission design,'' in \emph{Equadiff 99: (In 2 Volumes)}.\hskip 1em plus 0.5em minus 0.4em\relax World Scientific, 9 2000, pp. 1167--1181.

\bibitem{cunningham2023interpolated}
D.~Cunningham and R.~P. Russell, ``An interpolated second-order relative motion model for gateway,'' \emph{The Journal of the Astronautical Sciences}, vol.~70, no.~4, p.~26, 2023.

\bibitem{kulik2023state}
J.~Kulik, W.~Clark, and D.~Savransky, ``State transition tensors for continuous-thrust control of three-body relative motion,'' \emph{Journal of Guidance, Control, and Dynamics}, pp. 1--10, 5 2023.

\bibitem{cunningham2024spice}
D.~A. Cunningham and R.~P. Russell, ``Relative motion solutions around an arbitrary {SPICE} kernel trajectory,'' in \emph{Astrodynamics Specialist Conference, 2024}, 2024, pp. AAS 24--457.

\bibitem{boone2024efficient}
S.~Boone and J.~McMahon, ``An efficient approximation of the second-order extended {K}alman filter for a class of nonlinear systems,'' in \emph{2024 European Control Conference (ECC)}.\hskip 1em plus 0.5em minus 0.4em\relax IEEE, 2024, pp. 3533--3538.

\bibitem{bernstein2009matrix}
D.~S. Bernstein, \emph{Matrix mathematics: theory, facts, and formulas}.\hskip 1em plus 0.5em minus 0.4em\relax Princeton university press, 2009.

\end{thebibliography}

\end{document}